\documentclass[a4paper, 11pt]{article}
\usepackage[utf8]{inputenc}
\usepackage[margin=2cm]{geometry} 

\usepackage{geometry}
\usepackage{amsmath}
\usepackage{amsfonts}
\usepackage{graphicx}
\usepackage{makecell}
\usepackage{textcomp}
\usepackage{subcaption}
\usepackage{arydshln}
\usepackage{hyperref}

\usepackage[font=footnotesize, labelfont=bf]{caption}

\usepackage{newfloat}
\DeclareFloatingEnvironment[name={Supplementary Figure}]{suppfigure}
\DeclareFloatingEnvironment[name={Supplementary Table}]{supptable}

\makeatletter
\def\blfootnote{\xdef\@thefnmark{}\@footnotetext}
\makeatother

\begin{document}
\begin{center}
    \Large \textbf{Modality Attention and Sampling Enables Deep Learning with Heterogeneous Marker Combinations in Fluorescence Microscopy}\\
    \vspace{.5em}
    \small Alvaro Gomariz$^{1,2,*}$, Tiziano Portenier$^1$, Patrick M. Helbling$^2$, Stephan Isringhausen$^2$, Ute Suessbier$^2$, \\ C\'esar Nombela-Arrieta$^{2,\dagger}$, Orcun Goksel$^{1,3,\dagger}$ \\ 
    \vspace{.5em}
    \footnotesize
    $^1$ Computer-assisted Applications in Medicine, Computer Vision Lab, ETH Zurich, Switzerland\\
    $^2$ Department of Medical Oncology and Hematology, University Hospital and University of Zurich, Switzerland\\
    $^3$ Department of Information Technology, Uppsala University, Sweden
    \blfootnote{$^*$ Corresponding author e-mail: alvaroeg@ethz.ch}
    \blfootnote{$^\dagger$ These two last authors have equal contributions}
\end{center}

\begin{abstract}
Fluorescence microscopy allows for a detailed inspection of cells, cellular networks, and anatomical landmarks by staining with a variety of carefully-selected markers visualized as color channels. 
Quantitative characterization of structures in acquired images often relies on automatic image analysis methods. 
Despite the success of deep learning methods in other vision applications, their potential for fluorescence image analysis remains underexploited. 
One reason lies in the considerable workload required to train accurate models, which are normally specific for a given combination of markers, and therefore applicable to a very restricted number of experimental settings.
We herein propose \emph{Marker Sampling and Excite} --- a neural network approach with a modality sampling strategy and a novel attention module that together enable ($i$)~flexible training with heterogeneous datasets with combinations of markers and ($ii$)~successful utility of learned models on arbitrary subsets of markers prospectively. 
We show that our single neural network solution performs comparably to an upper bound scenario where an ensemble of many networks is naïvely trained for each possible marker combination separately. 
In addition, we demonstrate the feasibility of this framework in high-throughput biological analysis by revising a recent quantitative characterization of bone marrow vasculature in 3D confocal microscopy datasets and further confirm the validity of our approach on an additional, significantly different dataset of microvessels in fetal liver tissues. 
Not only can our work substantially ameliorate the use of deep learning in fluorescence microscopy analysis, but it can also be utilized in other fields with incomplete data acquisitions and missing modalities.
\end{abstract}

\section*{Introduction}
Deep neural networks have been largely successful in many computer vision problems~\cite{lecun2015deep,litjens2017survey}, by learning network model parameters in layers to produce feature maps called \emph{activations}, to arrive at a desired output often given as ground truth during a \emph{training} phase.
The learned parameters can then be deployed in an \emph{inference} phase to make predictions on new input data. 
Countless advances in this field have occurred in a relatively short timeframe, especially in the use of supervised segmentation, where the goal is the semantic partitioning of an input image with the ground truth typically consisting of pixels annotated interactively by experts.
For instance, \emph{UNet}~\cite{ronneberger2015u} is a well-known deep Convolutional Neural Network (CNN) architecture with proven success on semantic segmentation in various biomedical domains. 
Nevertheless, some aspects of biological images still pose several practical challenges in the application of deep CNN architectures (hereafter also called \emph{models}).

Fluorescence-based microscopy (FM) is  a mainstay technology for the study of living or fixed tissues in biomedical research. It operates by detecting microscopic signals emanating from inorganic molecules or genetically encoded proteins. 
Fluorescent dyes are often coupled to antibodies, which target structures or cells of interest within complex samples in a highly specific fashion, a process known as \emph{immunostaining}. 
Fluorescent signals are registered and separately encoded as independent image \emph{channels} due to their distinct spectral properties, thereby allowing the visualization of stained anatomical landmarks of interest. 
Herein we refer to these channels as \emph{markers} (also called \emph{labels} in the literature), which are analogous to the acquisition of \emph{modalities} in other imaging fields, such as the specific imaging sequences for quantifying different tissue properties with Ultrasound or Magnetic Resonance Imaging. 

The inherent nature of markers in bioimaging studies poses some additional limitations in the creation of datasets that can be processed by typical CNN frameworks. 
First, the number of markers that can be simultaneously imaged is limited, due to the overlapping spectral profiles of different fluorochromes, which preclude their reliable separation in individual channels.
Therefore, any detailed characterization of tissues and their pathological perturbations often requires the use of different permutations of a restricted number of markers, which in turn can only provide a limited level of insight into the biological structures studied. 
Moreover, sample availability is typically a limiting factor, and processing, immunostaining and image acquisition are laborious and time consuming tasks, especially for whole-organ or 3D imaging techniques. 
Thus, it is not always technically feasible to increase the number of markers, although such additional sources of information would simplify image processing techniques.
Finally, the process of immunostaining does not always work  consistently, leading to cells and structures stained with variable intensity despite using the same markers.
Altogether, these issues hinder the generation of a large number of datasets of images stained consistently with all combinations of possible markers. 
FM datasets thereby often consist of heterogeneous combinations of markers, and with each combination often being limited in number of samples, applications of deep learning become strongly limited. 
Furthermore, typical supervised segmentation algorithms allow a trained model for only limited future applicability, i.e.\ when the exact same marker combination is used as in training. 
This limitation leads to the tradeoff that either a separate specific model is trained each time a new combination is desired and data is available, or a small set of intersection of markers is found in the data; either way neglecting large amounts of precious data and any possibility of using the models later with alternative marker combinations. 

In a general image analysis framework, the problem settings above can be referred to as \emph{missing modalities}, and are somewhat related to Multi-Task Learning, a field that studies whether information should be learnt jointly or separately \cite{zhang2018overview}, and to Domain Adaptation, which aims to bring datasets from different sources into a common space to improve generalization performance~\cite{wang2018deep}. 
It is agreed in both these fields that using a unique model that shares certain amount of information is advantageous. 
Despite many advances in these fields, the presented missing marker or modality problem is, however, largely unexplored. 
Recently, synthesis approaches for completing missing data have been proposed for both markers~\cite{guo2020revealing, ounkomol2018label, christiansen2018silico} and time-sequences~\cite{ouyang2018deep} in FM, as well as for modalities in magnetic resonance imaging~\cite{li2014deep, iglesias2013synthesizing, chartsias2017multimodal, lee2019collagan}. 
Such modality synthesis is cumbersome and potentially sub-optimal when the segmentation model could instead encode information across modalities with shared features. 
Methods combining different modalities in shared feature spaces were proposed in \cite{havaei2016hemis} as Hetero-Modal Image Segmentation framework (\emph{HeMIS}) as well as in \cite{dorent2019hetero, varsavsky2018pimms}.
One would reasonably expect a multi-modal network model to behave differently in the existence or absence of a particular modality.
Such processes of conditioning the models explicitly are known as \emph{attention} mechanisms.
For example, soft attention mechanisms transform the activations of a model conditioned on the activations themselves \cite{jaderberg2015spatial, wang2017residual}.  
Notably, \emph{Squeeze and Excitation} (\emph{SE}) \cite{hu2018squeeze} and similar modules \cite{roy2020squeeze, roy2018recalibrating, rickmann2019project, wang2019towards} have been very successful and since been integrated in several different network architectures to improve their performance and the interpretability of extracted features. 

Building on these ideas, we herein devise a method that addresses the fundamental problem of multi-modality heterogeneous sets and evaluate this on a 3D microscopy image dataset of bone marrow vascular network \cite{gomariz2018quantitative}, where the annotations are divided into two vascular types, namely \emph{sinusoids} and \emph{arteries}. 
The variable size and morphology of the vasculature had hindered precise segmentation and thus a reliable vasculature characterization in earlier works \cite{gomariz2019imaging, acar2015deep, coutu2017three, coutu2018multicolor, christodoulou2020live}.
In this experimental setting with 5 FM imaging markers, we first evaluate multiple baseline conventional scenarios to analyze the effect of each possible marker combination on the performance of semantic (not to be confused with instance) vessel segmentation, which also serves as an upper bound for assessing our proposed methods. 
Then, we show that a \emph{Marker Sampling} strategy enables a single CNN model to successfully perform in the presence of any marker combination, while outperforming the current state-of-the-art \emph{HeMIS}. 
Next, we present our novel \emph{Marker Excite} soft attention module, which learns how to recalibrate the network activations as a function of available markers, showcasing this by training a \emph{single} model that performs comparably to the upper bound scenario with an ensemble of 31 separate models individually trained and specialized for each of the marker combinations. 
We further demonstrate that our model can even \emph{outperform the upper bound}, by leveraging information shared across markers when the training dataset contains such practical variations. 
Next, we present a case study  on the problem setting of~\cite{gomariz2018quantitative} to show the application of our method on an existing practical research question. 
Finally, we demonstrate the widespread applicability of our methods by applying them to a largely different and independently generated fetal liver FM dataset, thereby providing the first quantitative characterization of microvascular networks in this organ across different embryonic stages.  

\section*{Results}
\subsection*{Segmenting fluorescence microscopy samples stained with multiple markers}\label{sec:baseline}

\begin{figure}[p]
    \centering
    \includegraphics[width=0.83\textwidth]{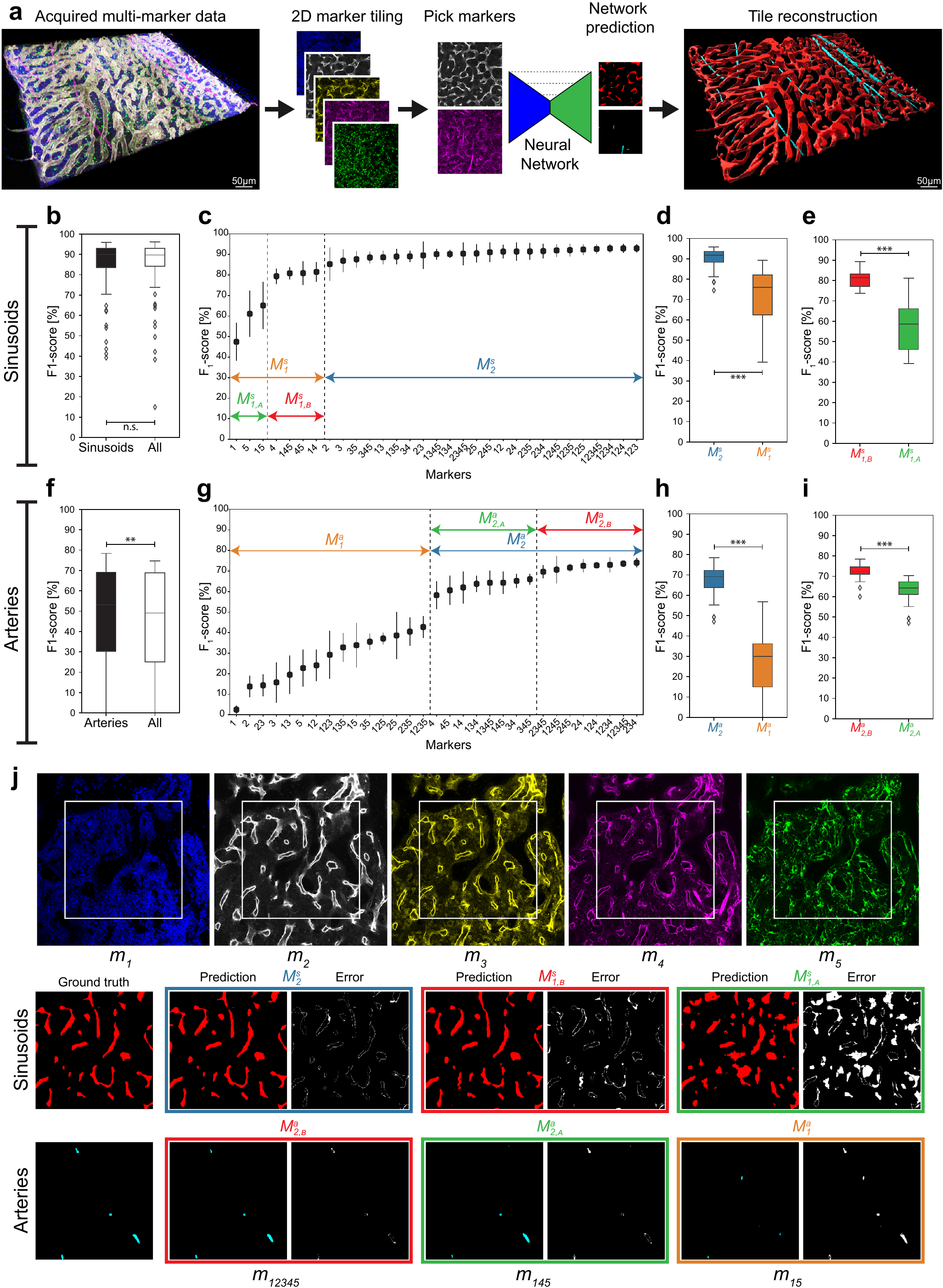}
    \caption{
    Segmentation of different classes with different \emph{UNet} models trained with specific marker combinations. 
    \textbf{a}~Illustration of our image segmentation pipeline in the presence of multiple markers. 
    \textbf{b,f}~Comparison of $F1$-score (n=124) evaluated separately on sinusoids (\textbf{b}) and arteries (\textbf{f}) for all CNN models (across marker sets and cross-validations) when trained separately to segment the class under evaluation as well as when trained to segment both classes simultaneously (\emph{All}). 
    \textbf{c,g}~$F1$-score (mean $\pm$ standard deviation) for models trained with each possible combination of markers, sorted in ascending order of mean values. 
    Each model was evaluated through 4-fold cross-validation (n=4). 
    We group markers as $M_i^c$ based on which models perform similarly on class $c$ ($c=s$ for sinusoids and $c=a$ for arteries). 
    The subscript $i$ denotes different groups. 
    \textbf{d}~For sinusoids, models with either $m_2$ or $m_3$ marker ($M^s_2$, n=96) are seen to perform superior to those without these markers ($M^s_1$, n=28). \textbf{e}~Among the remaining $M^s_1$ models, those with $m_4$ ($M^s_{1,B}$, n=48) perform superior to those without ($M^s_{1,A}$, n=48).
    \textbf{h}~For arteries, models with $m_4$ ($M^a_2$, n=64) perform superior to those without $m_4$ ($M^a_1$, n=60), and 
    \textbf{i}~among $M^a_2$, the ones with $m_2$ ($M^a_{2,B}$, n=32) perform superior to those without ($M^a_{2,A}$, n=32).
    \textbf{j}~Visualization of the different markers employed (upper row) and the ground truth and segmentation predictions for models using sample marker combinations as examples of different aforementioned groups. 
    The white squares within the marker images depict the size of their corresponding segmentation images.
    Error figures show false positive and false negative pixels.
    Significance is indicated with p-value$\leq$0.05~(*), p-value$\leq$0.01~(**), and p-value$\leq$0.001~(***). 
    }
    \label{fig:f1}
\end{figure}

Markers in FM label specific biological structures, and their efficient combination enables the visualization of distinct cellular/subcellular components or networks thereof.
Manual annotation of these structures as classes is usually possible when the available markers accurately portray them. 
Meanwhile, image segmentation algorithms can target these classes by employing arbitrary combinations of markers, but their performance will largely vary depending on the combination employed. 
To be able to achieve such segmentation with different markers, we designed a neural network-based image processing pipeline applicable to tissue-wide FM imaging (illustrated in Fig.\,~\ref{fig:f1}a with details in the Methods section) that processes markers as channels. 
In this work we employ a dataset with 8 large samples divided in a number of 2D patches (Supplementary Table~\ref{tab:dataset_bm}). 
Each patch consists of two ground truth classes (sinusoids and arteries) and five markers (DAPI, endomucin, endoglin, collagen, and CXCL12-GFP) denoted as $m_G, \; G \subseteq \{1, \dots, 5\}$. 
Hereafter, we use multiple subscripts successively to indicate combinations of these markers. 
All following CNNs results were validated with 4-fold cross-validation, and  claims are made only when the respective null hypothesis can be rejected with a p-value$\leq$0.05. 
More details are given in the Methods Section. 

We build an extensive semantic segmentation baseline by training a distinct \emph{UNet} separately for each of the 31 possible combinations of 5 markers.
These also serve in the following sections as upper bound performance given this architecture.
We present results for models trained separately for two classes of vessels, since empirically (measured with $F1$-score in Fig.\,~\ref{fig:f1}b,f) this performed  superior to training them simultaneously. 
To analyze marker importance, we rank in Fig.\,~\ref{fig:f1}c-e\&g-i the marker combinations according to their mean $F1$-score.
Although the segmentation accuracy is seen overall to increase with more markers, it is seen to also highly depend on specific markers: 
Sinusoid segmentation in Fig.\,~1c-e has higher accuracy when either $m_2$ or $m_3$, which specifically label these structures, are present (blue). 
Without them, $m_4$ helps with segmentation (red), compared to the least helpful $m_1$ or $m_5$ (green). 
For arteries in Fig.\,~1g-i, having the arterial-specific $m_4$ marker seems essential (blue), and results can be further improved by adding $m_2$ (red). 
These observations are used later herein to interpret outcomes, e.g.\ when some suitable markers are missing.

\subsection*{\emph{Marker Sampling} for segmenting with missing markers}\label{sec:ms}

\begin{figure}[b!]
    \centering
    \includegraphics[width=0.85\textwidth]{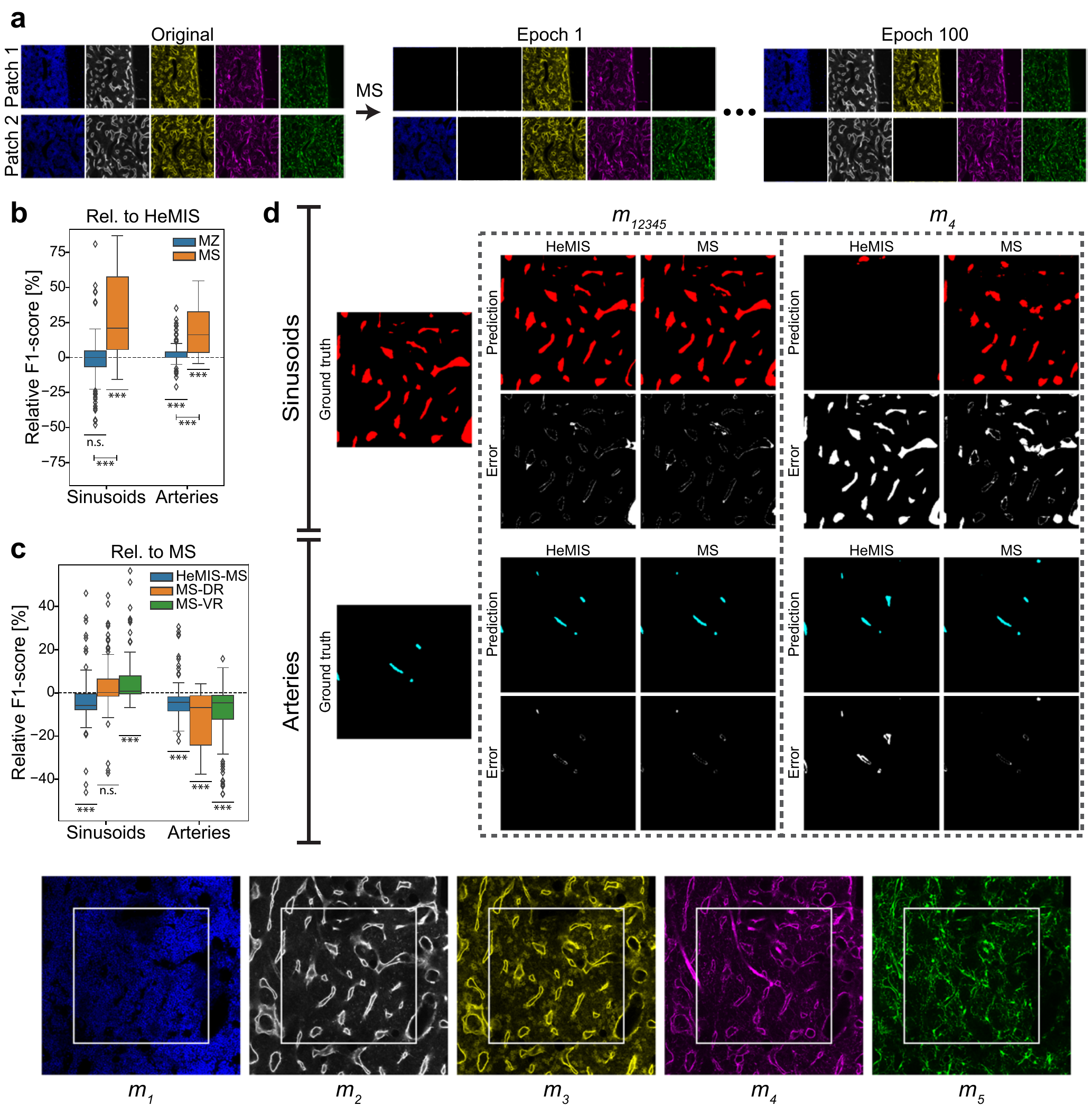}
    \caption{
    Segmentation with a single CNN model for all marker combinations. 
    \textbf{a}~Illustration of the \emph{MS}~strategy. Every time a batch is fed to the network during training, its markers are randomly deleted, i.e.\ set to blank (zero) images. 
    \textbf{b}~Comparison of \emph{MZ} and \emph{MS}~with \emph{HeMIS}. To this end, a paired test is employed to compare $F1$-score of each marker set between the model and HeMIS, representing all differences (n=124) as a boxplot for sinusoids and arteries.
    \textbf{c}~The same representation of results (n=124) is employed to compare \emph{HeMIS-MS}, \emph{MS-DR}, and \emph{MS-VR} with respect to \emph{MS}. 
    \textbf{d}~Visual examples of the differences in segmentation with \emph{HeMIS} and \emph{MS} for different marker combinations (shown on bottom). 
    When all five markers are used, no significant differences are observed.
    But when a subset is used (for instance, $m_4$ alone), the proposed \emph{MS} performs much superior to \emph{HeMIS}.
    Significance is indicated with respect to the baseline model of each graph (-----) or between different models ($|$\!-----\!$|$) with p-value$\leq$0.05~(*), p-value$\leq$0.01~(**), and p-value$\leq$0.001~(***).  
    }
    \label{fig:f2}
\end{figure}

The strategy described above allows to  determine the best combination of markers to segment a given class.
Nevertheless, the need for a fixed marker combination would be restraining in different applications due to the practical limitation on number of markers to use simultaneously in sample preparation. 
It is also not fail-safe if individual markers fail during acquisition. 
Moreover, such a naïve approach of training a distinct CNN for each combination would require $2^K - 1$ models for $K$ markers (31 models in our study), an exponential growth, which is prohibitive for training CNNs in reasonable time frames.
Finally, retraining this many models becomes highly impractical as new samples are added to the training dataset. 
To address this challenge we propose a \emph{Marker Sampling} (\emph{MS}) approach (illustrated in Fig.\,~\ref{fig:f2}a). 
A key component of \emph{MS} is a sampling layer at the input of a segmentation network, herein \emph{UNet}. 
During training, we provide all the available markers to the network, while this sampling layer randomly selects a subset to be processed by the proceeding network. 
This single-model framework consequently can learn to generalize to any subset of markers at inference.

Since the accuracy for each marker combination may vary widely (also seen in Fig.\,~\ref{fig:f1}c,g), we herein report comparative improvements in a paired test manner; i.e.\ 
when we test a hypothesis with respect to an alternative method (indicated in the figures as ``Rel.\ to''), we calculate the relative metric difference separately for each marker combination experiment and report the distribution and statistical significance of such differences (with details in the Methods Section).
Fig.\,\ref{fig:f2}b shows a comparison of our \emph{MS} method with the state-of-the-art \emph{HeMIS}.
In addition we train our single CNN without \emph{MS}, i.e., the network always uses all markers during training without sampling. 
When performing inference on a subset of markers, we simply set the missing input to zero. 
We refer to this simplistic baseline as \emph{Marker Zero (MZ)}.
The results show that our \emph{MS} model vastly outperforms both \emph{HeMIS} and \emph{MZ}, indicating that training a network simply with randomly sampled marker subsets generalizes across the possible input combinations better than the other two approaches. 
Surprisingly, \emph{MZ} does not perform significantly different than \emph{HeMIS}, suggesting that the latter does not provide any advantage for this task compared to a standard \emph{UNet} architecture. 
Based on this, learning shared features can be considered comparable to marker-specific representations for this purpose.
For completeness we also evaluate incorporating the \emph{MS} strategy in the \emph{HeMIS} architecture (Fig.\,~\ref{fig:f2}c) and find that there is no improvement compared to \emph{UNet}-based \emph{MS}, therefore concluding that any advantage originates from our \emph{MS} strategy per se.

It could be beneficial to normalize the output of the sampling layer across channels to keep its signal magnitude constant. 
In fact, \emph{MS} can be interpreted as a variant of Dropout regularization of neural networks~\cite{srivastava2014dropout}, where it is applied only on the input layer and, not only during training but also during inference (deterministically via available markers). 
We investigate such relation by training two variants of our model: \emph{Marker Sampling with Dropout} (\emph{MS-DR}), which scales the intensities of the available images by a constant Dropout ratio, and \emph{Marker Sampling with Variable Ratio} (\emph{MS-VR}), which scales the intensities by the ratio of available markers in each sample. 
Among these, \emph{MS} is found to be the best performing model (Fig.\,~\ref{fig:f2}c), indicating that our improvements cannot be matched by such handcrafted normalization in the sampling layer. 

In summary, while \emph{HeMIS}, \emph{MZ}, and \emph{MS} all perform well when all markers are present ($m_{12345}$), our proposed \emph{MS} generalizes significantly better when some markers are missing at the time of segmentation. 
Indeed, for instance, with $m_4$ alone our proposed \emph{MS} can already segment satisfactorily (Fig.\,~\ref{fig:f2}d), which the baselines are unable to.

\subsection*{\emph{Marker Excite} for learning attention to markers}\label{sec:msme}

\begin{figure}[p]
    \centering
    \includegraphics[width=0.85\textwidth]{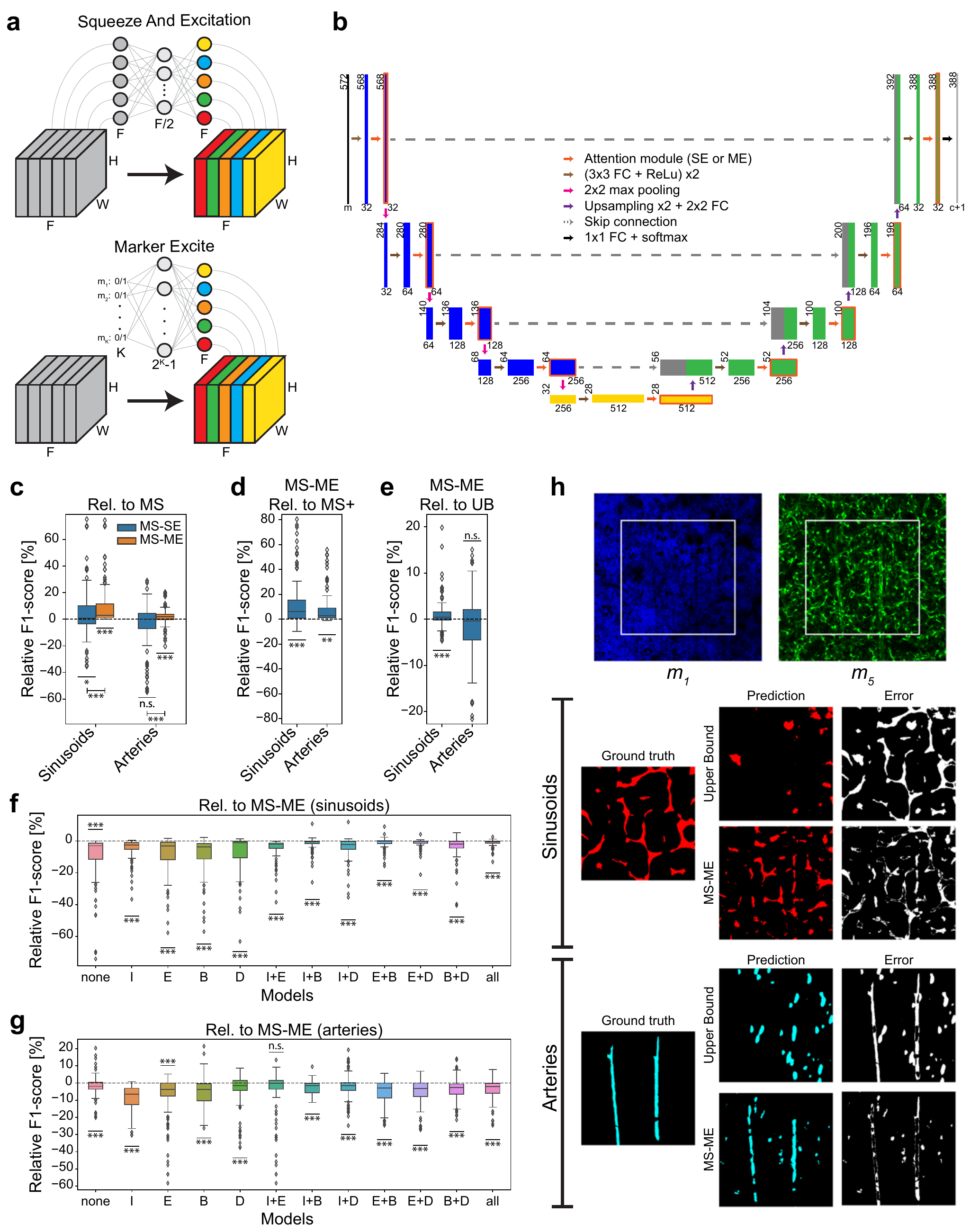}
    \caption{
    Results with attention modules. 
    \textbf{a}~Illustration of the attention modules: conventional \emph{SE} (top) and proposed \emph{ME} (bottom).
    These modules learn weights (colored circles) which are employed to recalibrate the activations (colored stacks) of a feature map (in gray). 
    Whereas \emph{SE} learns such weights as a function of spatially pooled feature maps, our proposed \emph{ME} learns from a symbolic one-hot encoded vector indicating marker availability (details in Methods). 
    \textbf{b}~Schematics of our network architecture based on \emph{UNet}, where the feature maps calibrated by an attention module are outlined in orange. 
    The numbers below each block indicate the number of activations in that feature map, and the numbers to their left specify their width and height.
    For ablation experiments, we refer to parts of the network as encoder (blue), bottleneck (yellow), decoder (green), and input (black). 
    \textbf{c}~Relative $F1$-score (n=124) of \emph{MS-SE} and \emph{MS-ME} compared pair-wise to \emph{MS}.
    Similarly, \emph{MS-ME} compared to \emph{MS+} in \textbf{d} and to UB in \textbf{e}.
    \textbf{f,g}~An ablation study for placing \emph{ME} attention at different network layers colored in \textbf{b}: input (I), encoder (E), bottleneck (B) and decoder (D). 
    Results for \textbf{f}~sinusoids and \textbf{g}~arteries are presented relative to the proposed \emph{MS-ME} (equivalent to \mbox{E+B+D)}.
    \textbf{h}~Visualization of a sample output with marker combination $m_{15}$ (among the worst according to Fig.\,~\ref{fig:f1}c,g), where it is seen that our multi-task single model \emph{MS-ME} performs superior to the dedicated upper bound \emph{UB} in $F1$-score, thanks to the \emph{ME} attention module and by potentially leveraging additional info from other marker combinations in the shared network body.  
    Significance is indicated with respect to the baseline model of each graph (-----) or between different models ($|$\!-----\!$|$) with p-value$\leq$0.05~(*), p-value$\leq$0.01~(**), and p-value$\leq$0.001~(***). 
    }
    \label{fig:f3}
\end{figure}

As demonstrated above, \emph{MS} can effectively generalize across marker combinations using a traditional neural network backbone such as \emph{UNet} to learn shared features for distinct markers. 
We herein additionally propose \emph{Marker Excite} (\emph{ME}) --- a novel attention module for a further boost by learning a weighting of deep features as a \emph{function of available markers}.  
Note that existing attention modules such as \emph{SE} learn such weighting as a function of layer activations.
Instead, our \emph{ME} module learns attention to marker availability, provided explicitly as an additional input to the network in the form of a \emph{one-hot encoded} symbolic vector (Fig.\,~\ref{fig:f3}a). 
We integrate \emph{ME} modules at different layers of a \emph{UNet} as shown in Fig.\,~\ref{fig:f3}b.
Integrating our \emph{ME} approach with the previously presented \emph{MS} strategy of prepending a sampling layer, we attain our ultimate model~\emph{MS-ME}.
We compare this below also using the conventional attention by replacing \emph{ME} with \emph{SE}, a baseline referred hereafter as \emph{MS-SE}.

Fig.\,~\ref{fig:f3}c shows that \emph{MS-ME} yields improved overall accuracy with respect to \emph{MS} as well as \emph{MS-SE}, especially for the sinusoids. 
A major advantage of attention modules is that they increase the model complexity only marginally (\emph{MS-ME} having merely 0.64\% more parameters than the \emph{UNet} architecture it was based on). 
To demonstrate that the presented improvement does not originate from inflation of model size, we conducted an additional experiment with a much larger \emph{UNet} baseline (\emph{MS+}) of over 20\% more parameters and show that the clear improvement from our proposed approach persists (Fig.\,~\ref{fig:f3}d). 
We also performed an ablation study by placing \emph{ME} attention at different network layers (Fig.\,~\ref{fig:f3}f,g), concluding that placing \emph{ME} after every convolutional block (outlined in orange in Fig.\,~\ref{fig:f3}b) is the optimal configuration (namely \emph{MS-ME}) for our task.
An analysis of the influence of the proposed \emph{ME} modules reveals that the recalibration effect is stronger for activations of higher resolution (Supplementary Fig.\,~\ref{fig:sf6}).

We also compare our proposed \emph{MS-ME} model with an upper-bound (\emph{UB}) presented in Fig.\,~\ref{fig:f1}c,g.
Note that \emph{UB} consists of 31 individual models each of which separately trained and specialized to the availability of a specific marker combination, whereas \emph{MS-ME} is a single model aiming to achieve well on whatever marker combination may be available at a given time.
Fig.\,~\ref{fig:f3}e shows that across all markers combinations our \emph{MS-ME} is not significantly different than \emph{UB}, and even slightly superior for sinusoid segmentation potentially thanks to leveraging additional information shared across combinations. 
A sample qualitative result is provided in Fig.\,~\ref{fig:f3}h for a marker combination ($m_{15}$) known to be suboptimal (based on Fig.\,~\ref{fig:f1}c,g) with a specialized training of a dedicated \emph{UB} model.
Even for this case, \emph{MS-ME} is seen to still perform somewhat satisfactorily, while \emph{UB} fails completely.

\subsection*{Training with heterogeneous panels of markers}\label{sec:tsb}
\begin{figure}[t]
    \centering
    \includegraphics[width=\textwidth]{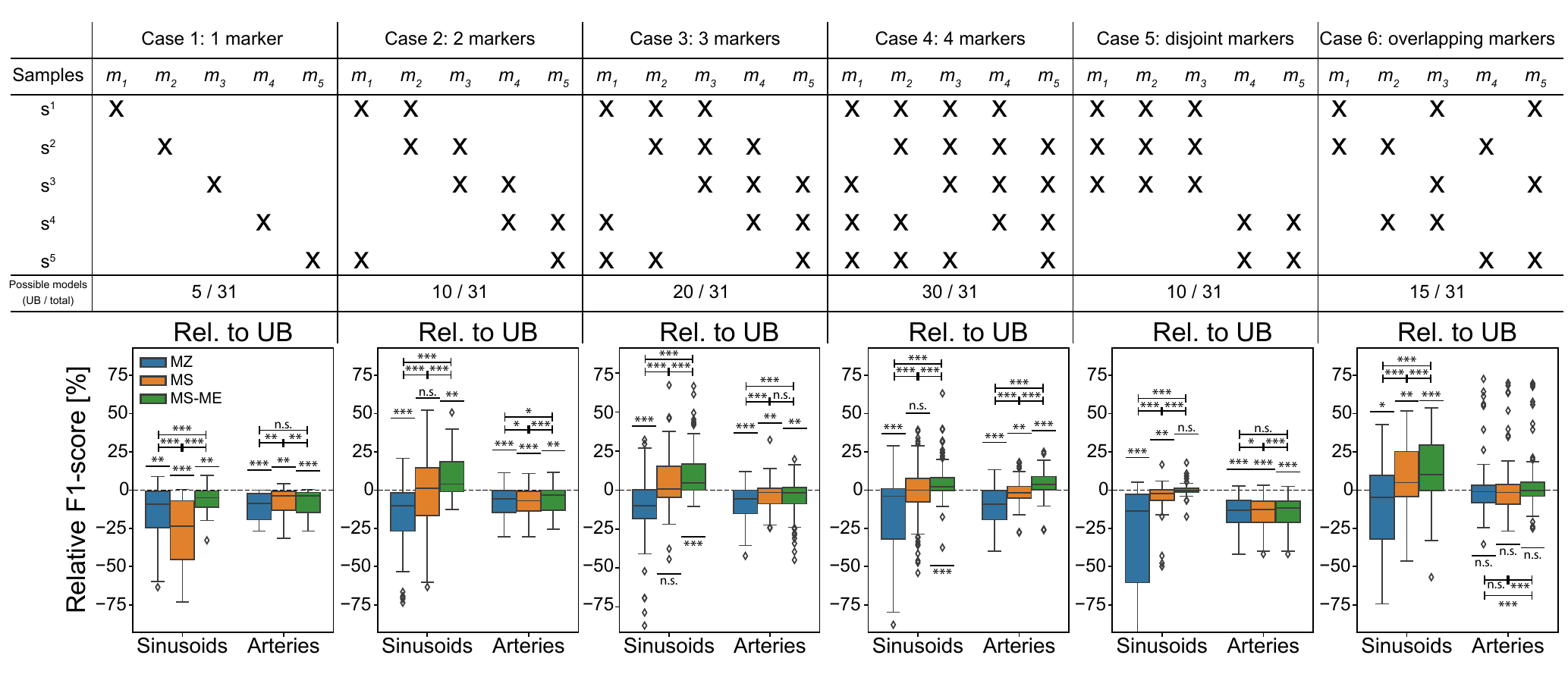}
    \caption{
    Comparison of proposed CNN models to UB when training with heterogeneous combinations of markers. 
    The upper table represents the markers available (denoted with X) for each of the 5 training samples in the different simulated cases. The number of marker combinations for which an UB model can be created as compared to our models (as explained in the text) in each Case is shown as an additional row (possible models). Below, the three most representative models proposed in this work (\emph{MZ}, \emph{MS} and \emph{MS-ME}) are compared for the segmentation of sinusoids and arteries in the 6 different Cases. 
    Significance is indicated with respect to the baseline model of each graph (-----) or between different models ($|$\!-----\!$|$) with p-value$\leq$0.05~(*), p-value$\leq$0.01~(**), and p-value$\leq$0.001~(***). 
    }
    \label{fig:f4}
\end{figure}

We next study the scenario with an incomplete training set, i.e.\ with acquisitions of heterogeneous combinations and different number of markers --- a typical setting in the field in practice.
For this purpose, we used subsets of our fully-annotated dataset.
Training a separate network of all combinatorial test settings would be computationally prohibitive, thus we artificially ablate the data to create a number of case studies (Fig.\,~\ref{fig:f4}) and thereby emulate various practical or extreme scenarios. 

In Cases 1 to 4 we studied settings in which training data contain a fixed number of markers per sample, each with a different marker combination. 
Case~5 simulates a common scenario where two different sets of samples are prepared with two different staining protocols.
Case~6 studies a practical scenario with samples available from different studies, which have so far been largely unusable for machine learning due to their heterogeneity.

While our proposed CNN models are by construction applicable for the above scenarios, any baseline segmentation methods such as \emph{UNet} require an overlapping set of markers. These baseline models are therefore trainable only for a few intersections of marker combinations, and discard any potentially useful information outside such common intersections.
For instance, the marker combination $m_{23}$ in Case~3 is only trainable using samples s1 or s2, and in Case~1 not even trainable.

With our proposed models, in all cases any marker combination is trainable. 
Therefore, for numerous test combinations, our proposed models are inherently superior by design, since the baselines cannot even accommodate test combos unseen during training.
For many other tests, the baselines would be using a small intersection subset of samples, again at a major disadvantage and also forcing us to retrain a network for each of these subsets.
We herein compared our methods (Fig.\,~\ref{fig:f4}) to the \emph{UB} models only for marker combinations achievable by baseline in training. 
As this clearly biases results towards \emph{UB}, we also report herein the ratio of marker combinations subject to such evaluation, i.e.\ available in the training set.
Small ratios in most cases demonstrate that our contribution is not only superior in $F1$-score, but also in the ability to include and utilize datasets hitherto impossible.

The superiority of \emph{MS} over \emph{MZ} and of \emph{MS-ME} over \emph{MS} is emphasized with increasing number of markers available for individual samples, since there are more combinations from which information can be leveraged through sampling and attention. 
Thus, \emph{MS} and \emph{MS-ME} do not add any advantage only when all samples have a single marker (Case~1). 
Remarkably, when considering both sinusoids and arteries, \emph{MS-ME} outperforms \emph{UB} in Cases 3, 4, and 6, where there is more heterogeneity of markers across samples. 
Thus, best relative gains of \emph{MS-ME} are observed with more heterogeneity and number of markers across samples. 
We see that \emph{MS-ME} not only requires a single model and is applicable for several marker combinations previously unattainable, but it also presents a superior segmentation performance, even over the \emph{UB} for several cases.

\subsection*{A practical application on segmenting bone marrow vasculature}\label{sec:appl_bm}
\begin{figure}[t]
    \centering
    \includegraphics[width=0.95\textwidth]{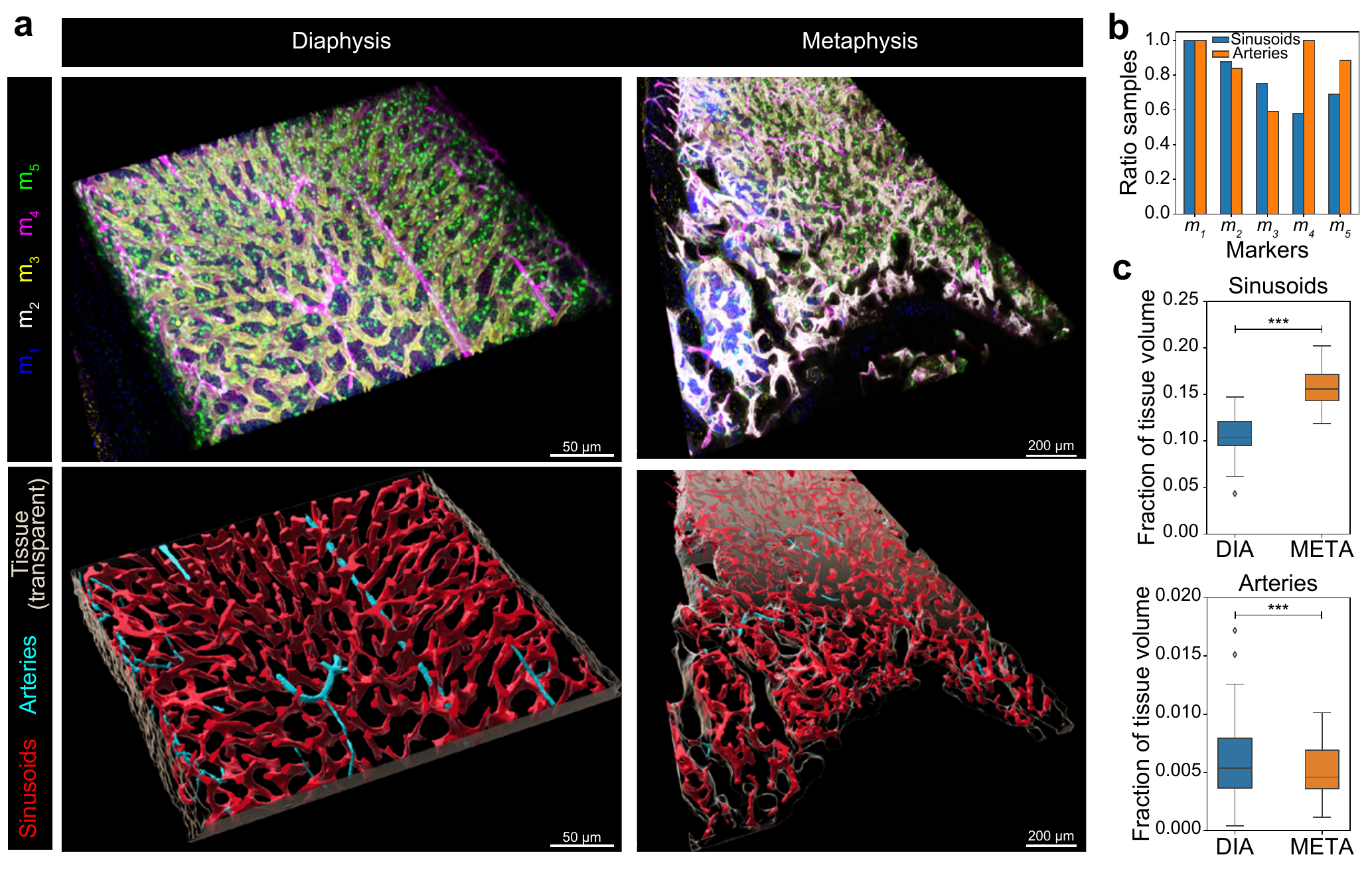}
    \caption{
    Vasculature segmentation and quantification in bone marrow with \emph{MS-ME}. 
    \textbf{a}~Visual example of bone marrow images for diaphysis, DIA (left) and metaphysis, META (right). Each sample has a marker combination (displayed as a maximum intensity projection in the top row) which is employed by our \emph{MS-ME} to segment arteries and sinusoids (bottom row). The tissue mask is shown as a transparent 3D mask. Image visualization with Imaris (Bitplane AG). 
    \textbf{b}~Fraction of quantified images that contained each of the markers for both sinusoids and arteries. The number of images is different due to the quality standards defined in the Methods Section.
    \textbf{c}~Fraction of bone marrow volume occupied by the different vascular structures for diaphysis (DIA, n=47 for sinusoids and n=29 for arteries) and metaphysis (META, n=14 for sinusoids and n=6 for arteries) regions. 
    Significance is indicated with p-value$\leq$0.05~(*), p-value$\leq$0.01~(**), and p-value$\leq$0.001~(***).
    }
    \label{fig:f5}
\end{figure}

As a showcase of the proposed methods in an application scenario, we herein revise our segmentation results from~\cite{gomariz2018quantitative}, where the use of traditional Morphological Image Processing (MIP) algorithms had restricted our analyses both in terms of the number of samples included as well as in the accuracy of quantification. 
As indicated in Table~\ref{tab:bmquant}, the herein proposed \emph{MS-ME} ($i$)~permits full automation of the analysis, ($ii$)~substantially increases $F1$-score by 47.3$\pm$9.6\%, and ($iii$)~almost quadruples the number of samples that can be successfully processed, allowing the inclusion of 35 samples which could not be previously analyzed due to marker heterogeneity (Fig.\,~\ref{fig:f5}b) or  insufficient image quality for MIP. 
Segmentation examples with \emph{MS-ME} can be seen in Supplementary Fig.\,~\ref{fig:sf1}.

\begin{table}[b]
    \centering
    \begin{tabular}{c|c:c|c:c}
        & \multicolumn{2}{c|}{ sinusoids } & \multicolumn{2}{c}{ arteries } \\
        \cline{2-5}
        & MIP \cite{gomariz2018quantitative} & MS-ME (proposed) & MIP \cite{gomariz2018quantitative} & MS-ME (proposed) \\
        \hline
        $F1$-score & 61.9$\pm$15.7 & 91.2$\pm$3.9 & n/a & 71.2$\pm$4.4 \\
        \# samples & 12 & 47 & n/a & 29 \\
        manual work & 30 min/sample & automatic & n/a & automatic \\
    \end{tabular}
    \caption{
    Comparison of bone marrow vasculature quantification as previously reported with segmentation based on Morphological Image Processing (MIP), and with the \emph{MS-ME} model proposed in the current work. Arteries appear as not available (n/a) because their segmentation could not be achieved with the earlier MIP approach. Manual work herein refers to the user interaction required to apply a method on a sample. 
    }
    \label{tab:bmquant}
\end{table}

We separately quantify the vascular volumes in two distinct regions of the bone marrow, namely \emph{diaphysis} and \emph{metaphysis}, shown in Fig.\,~\ref{fig:f5}a. 
Our revision of the previous  analysis~\cite{gomariz2018quantitative} indicates that the volume occupied by sinusoids is significantly lower than we previously reported (Supplementary Fig.\,~\ref{fig:sf2}).
The results in Fig.\,~\ref{fig:f5}c also confirm the previously observed trend (with more statistical power) that they are more abundant in the metaphysis ($15.65 \pm 2.55 \%$) than in the diaphysis ($10.69 \pm 2.16 \%$). 
The differences can be explained by the higher accuracy of our new method. 
In addition, our methods allow for an accurate segmentation of much rarer arterial networks, which to the best of our knowledge have only been quantified in very limited tissue regions in previous studies~\cite{Kunisaki_2013}. 
We find that these vessels occupy $0.63 \pm 0.40 \%$ of the volume in the diaphysis and $0.52 \pm 0.32 \%$ in the metaphysis.

\subsection*{Marker Sampling and Excite for the characterization of fetal liver vasculature}

\begin{figure}[hbt!]
    \centering
    \includegraphics[width=0.9\textwidth]{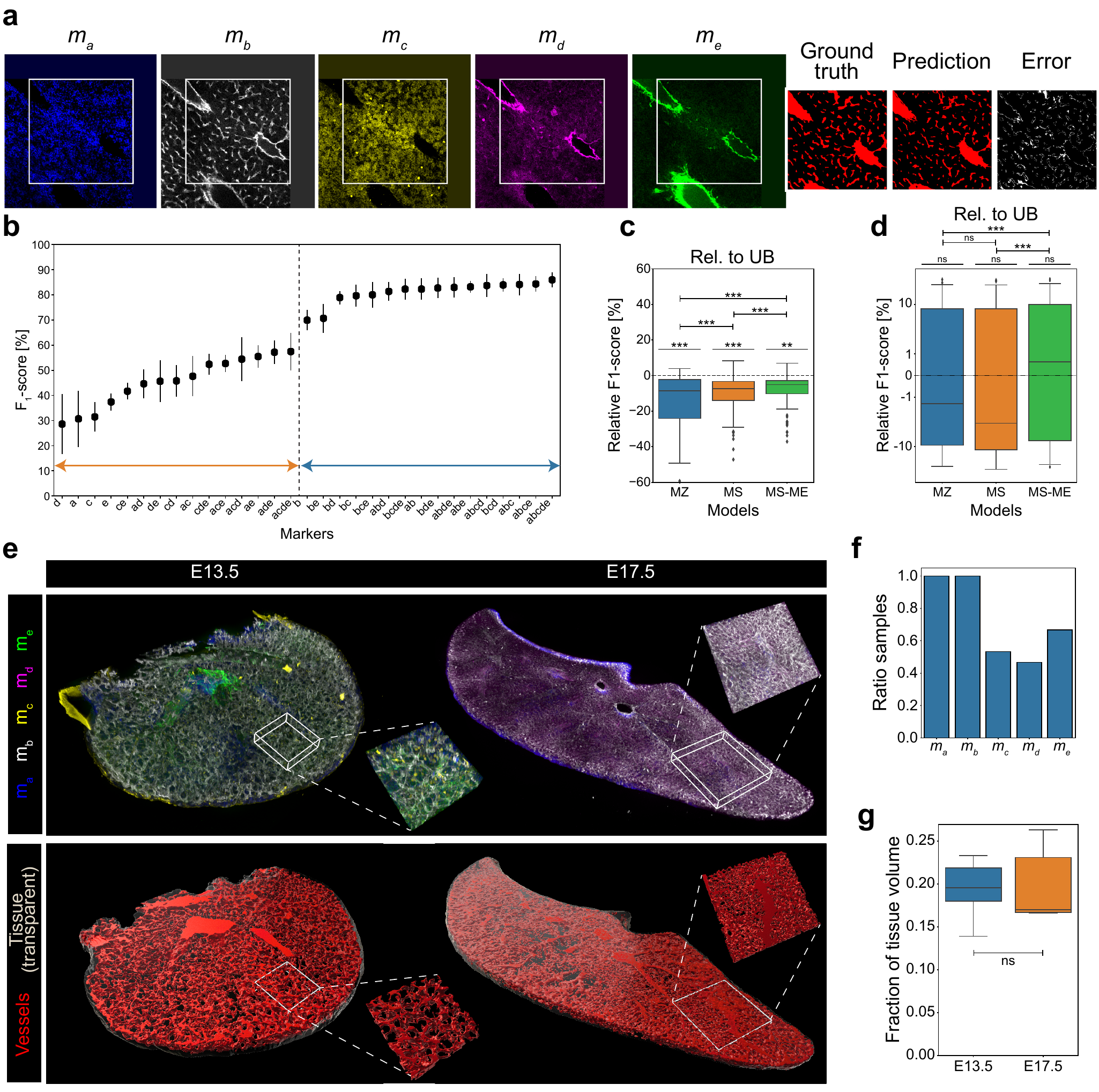}
    \caption{
    Study of our proposed models in the segmentation of fetal liver vasculature with different marker combinations. 
    \textbf{a}~Example of labeled patch that illustrates the markers available in this dataset with their corresponding segmentation output as ground truth, model prediction, and error. 
    \textbf{b}~$F1$-score achieved when training \emph{UB} models specialized on each of the possible marker combinations. 
    The segmentation quality is observed to be superior when $m_b$ is available (blue region) as compared to marker combinations without it (orange region).
    \textbf{c, d}~Segmentation $F1$-score of the different proposed models relative to \emph{UB} when (\textbf{c}) all markers are available at training time and (\textbf{d})~they are artificially ablated as described in the main text. 
    A symmetrical logarithmic scale (linear between -2 and 2, and logarithmic elsewhere) is used in \textbf{d} to emphasize the differences near the \emph{UB} line.
    \textbf{e}~Example visualization of two fetal liver samples corresponding to embryonic stages E13.5 (left) and E17.5 (right). 
    The maximum intensity projections of the volumes for the available markers are displayed in the top row and their corresponding vessels segmented with \emph{MS-ME} together with the tissue mask are shown below.
    \textbf{f}~Ratio of availability of the 5 different markers across the samples employed for quantification of vascular occupancy. 
    \textbf{g}~Fraction of tissue volume occupied by vasculature as segmented by \emph{MS-ME}, compared across fetal liver samples in the different embryonic stages E13.5 and E17.5. 
    Significance is indicated with respect to the baseline model of each graph (-----) or between different models ($|$\!-----\!$|$) with p-value$\leq$0.05~(*), p-value$\leq$0.01~(**), and p-value$\leq$0.001~(***). 
    }
    \label{fig:f6}
\end{figure}

To further demonstrate the ability to generalize across samples and the applicability of our proposed methods on significantly different biological tissues in which different marker combinations are available, we provide a comprehensive quantitative analysis of a murine fetal liver FM dataset.
The segmentation label (\emph{vessels}) and the 5 different markers employed for this dataset, largely divert from those employed in bone marrow and are illustrated in Fig.\,~\ref{fig:f6}a, denoted as $m_G, \; G \subseteq \{a, \dots, e\}$ (details in Methods section - Segmentation of fetal liver vasculature). 

Evaluating the \emph{UB} for each of the possible marker combinations shows that a superior $F1$-score is always achieved when $m_b$ is available (Fig.\,~\ref{fig:f6}b).
If all markers are available at training time, our models behave similarly as with the bone marrow dataset; i.e., \emph{MS} is superior to \emph{MZ} and \emph{MS-ME} is superior to \emph{MS} in terms of $F1$-score evaluated across all possible marker combinations at test time (Fig.\,~\ref{fig:f6}c). 
We further evaluate the segmentation performance when the training samples have different combinations of markers by artificially assigning a random marker combination to each of the samples. 
The results in Fig.\,~\ref{fig:f6}d show again the superiority in $F1$-score of \emph{MS-ME} relative to \emph{MS} and \emph{MZ}.
In this case, \emph{MZ} has a comparable $F1$-score to that of \emph{MS}, probably because the ablation strategy employed in this setting already creates enough training examples with the different combinations of markers, rendering sampling unnecessary. The $F1$-score of all three models is comparable to that of \emph{UB} in this setting, corroborating results from some training settings in Fig.\,~\ref{fig:f4} for the bone marrow dataset.

Next, we follow the analysis strategy of the previous section to demonstrate the potential of our framework in the characterization of different biological samples.
We employ \emph{MS-ME} to segment the liver vasculature in 15 fetal liver samples at two different embryonic stages denominated E13.5 and E17.5 as illustrated in Fig.\,~\ref{fig:f6}e, each having different marker combinations (Fig.\,~\ref{fig:f6}f).
The results in Fig.\,~\ref{fig:f6}g reveal isometric vascular growth during these embryonic stages, as no significant differences in vascular occupancy between timepoints is observed, with a vascular ratio of 19.4$\pm$3.60\% across all samples.  

\section*{Discussion}
In this work we formalize the widespread challenges concerning segmentation of structures in FM datasets with different markers, and address them with unprecedented accuracy and efficiency.
We first studied the contribution of all marker combinations by naïvely training a distinct CNN model for each combination. 
The segmentation results were ranked for each of the possible markers employed, of which even the worst combinations could produce meaningful results.
Albeit useful as an upper bound performance, training this number of models is infeasible in practice. 
We hence proposed a number of methods that allow for a practicable workflow with a single model that can operate with any combination of markers both during inference (application-time) and training.

During our preliminary experiments, the fact that a simple \emph{UNet} adaptation \emph{MZ} could perform as well as \emph{HeMIS}, the state-of-the-art deep learning approach for missing modalities, indicated to us that the solution to this challenge was potentially not in the design of novel network architectures, but rather in adapting suitable training strategies that can inherently address such problem. 
Accordingly, we devised \emph{MS} as a sampling strategy, the results of which confirmed our hypothesis, providing a drastic improvement at inference for segmenting different structures with any marker combination.
We also tested normalization strategies for the sampled markers, but none helped: \emph{MS-DR} may have failed because it scales image intensities by a constant factor at training time, which makes sense in the original Dropout idea that samples weights as a regularization technique, but sampling in our missing marker setting intrinsically occurs at inference too. 
\emph{MS-VR}, which scales intensities by the ratio of available markers, did not improve the results either. The reason could be that such scaling is reasonable as regularization in the case of marker-independent features, whereas the employed network may learn different features for each marker.

Results with our novel \emph{ME} method show that attention mechanisms do not have to be limited to layer activations, but using other sources of information as weak labels can also boost outcomes substantially with only a slight increase in model complexity.
When our two contributions were combined as \emph{MS-ME} the best segmentation results were obtained, performing as well as an ensemble of 31 specialized \emph{UB} models.
In addition, when training samples had different markers, \emph{MS-ME} was applicable to combinations inaccessible with typical CNNs such as \emph{UNet} or \emph{HeMIS}, while our \emph{MS-ME} segmentation accuracy surpassed \emph{even an upper bound setting}, especially when marker heterogeneity was high.
This surprising result, as a far-reaching consequence, shows that the solution we originally devised to accommodate missing markers can easily help to improve outcomes even in traditional learning scenarios, potentially by leveraging and incorporating complementary information from multiple sources. 

When applied to the bone marrow vasculature dataset, our approach based on \emph{MS-ME} showed a dramatic increase in accuracy, speed, and number of samples which could be processed, indicating the importance of chosen image analysis techniques in scientific studies.
We therewith increased the statistical power of the characterization reported in~\cite{gomariz2018quantitative}, allowing us to confirm the previously hinted tendency of sinusoids occupying larger volumes of the metaphysis than the diaphysis.
Application of the same analysis pipeline to a dataset depicting fetal liver vasculature with a different marker set, further allowed to quantitatively describe the vascular occupancy in different embryonic stages. 
Our proposed techniques will be instrumental in imaging studies to accurately study the morphometric and structural features of microvascular networks in hematopoiteic organs, to gain key deeper insight on their described fundamental roles in the pathophysiology of blood-forming tissues.
Beyond this, vascular segmentation in large multidimensional datasets remains a widespread and largely unresolved challenge for many groups using advanced microscopy technologies. The methods presented may reveal as extremely powerful tools for studies aimed at uncovering novel mechanisms regulating tissue and cellular dynamics in almost any biological tissue, from animal models, human samples, to model organoids grown in vitro~\cite{gomariz2019imaging}.

Despite the superiority of \emph{MS-ME} in segmentation results without compromises in number of parameter or inference speed (Supplementary Table~\ref{tab:complexity}), some limitations exist. 
For instance, although \emph{MS-ME} can be applied to previously unseen combinations of markers, quantitative quality assessment is only possible if labeled examples exist featuring the respective combination.
In addition, although not observed in our study, our \emph{MS} strategy may amplify existing imbalances of markers in the training set, increasing the danger of overfitting to specific combinations.

These challenges open a number of interesting questions for future work. 
For instance, novel uncertainty estimation methods for deep neural networks~\cite{gal2015cnn, kendall2017uncertainties} can be employed as proxy for assessment of segmentation quality upon previously unseen combinations. 
In addition, although overfitting to specific markers can be na\"ively addressed with a validation set containing as many markers as possible, investigating how to adapt the dropout rate of \emph{MS} to different experimental settings may lead to solutions with superior results.
Lastly, our \emph{MS-ME} model may enable novel possibilities in the context of transfer learning~\cite{transfusion_bengion19} that enables seamless application of trained networks to distinct biological tissues with new markers. 
For example, fine-tuning only the few parameters in the \emph{ME} module as opposed to expensive, classical fine-tuning strategies may yield competitive results if the core network already learned suitable general purpose features.

A major challenge and limitation for utilizing deep learning in FM has been the difficulty in establishing standardized staining protocols that would enable more homogeneous marker combinations to train supervised models.
With our methods proposed herein, a \emph{single} model is shown to perform comparably or superior to a number of individual problem-specific models that would be infeasible in practice due to the exponential growth in model parameters and training time with an increasing number of markers. 
In addition, the versatility of our methods enables them to be easily applied to different network architectures for tasks beyond semantic segmentation, such as classification~\cite{AlexNet, He2016}, detection~\cite{Xie2018, Ren2017}, or instance segmentation~\cite{He2017, Yang2020}.
These contributions can facilitate the sharing and exchange of trained CNNs across labs in the field as well as a faster adoption of neural network solutions in routine lab work at, e.g., microscopy facilities.

\section*{Methods}
\subsection*{Dataset composition and notation}
The dataset $S$ under evaluation consists of 8 samples $s^i, \, i \in \{1, \dots, 8\}$ and each sample $s^i$ is composed of $J$ patches  $p^i_j, \, j \in \{1, \dots, J\}$. 
Samples are prepared with a set of different markers $k \in \{1, \dots, K\}$, where $K$ is the total number of markers. Marker combinations are denoted as $m_G$, where $G$ is a non-empty subset of available markers, i.e.\ $G \subseteq \{1, \dots, K\}$, denoted in the subscript as a successive sequence, e.g.\ $m_{12345}$ indicating the combination of all markers for $K=5$ in our study. 
Each sample $s_i$ was manually annotated for $C$ classes $c \in \{1, \dots, C\}$. Thus, each patch $p^i_j$ consists of $\vert G \vert$ input images $x_k \in \mathbb{R}^{h_\text{in} \times w_\text{in}}$ with $k \in G$ and two output annotations $y_c \in \mathbb{R}^{h_\text{out} \times w_\text{out}}$ as follows: 
\begin{equation*}\label{eq:patchesmarkers}
 p^i_j = \{x_k, y_c\} \; \forall k \in G, \; \forall c \in  \{1, \dots, C\}   
\end{equation*}
All patches are of the same size, i.e.\ $h_\text{in} = w_\text{in} = 572$ and $h_\text{out} = w_\text{out} = 388$.

The complete bone marrow dataset $S$ (Supplementary Table~\ref{tab:dataset_bm}) was prepared using the following 5 markers: \emph{DAPI} stains DNA in all nuclei, \emph{endomucin} and \emph{endoglin} have both been reported to have a high specificity for bone marrow sinusoids, \emph{collagen} mostly stains vessel walls (including sinusoids and arteries), as well as extracellular matrix, and \emph{GFP} (in the CXCL12-GFP mouse) is a genetically encoded marker which stains reticular mesenchymal stromal cells. 
We use $C=2$ annotated classes: \emph{sinusoids} and \emph{arteries} (the latter accounting for both central larger arteries and endosteally located smaller arterioles). 
The immunostaining and imaging protocols are detailed extensively in~\cite{gomariz2018quantitative}.
Note that such a dataset with a fixed combination of markers and classes is challenging to obtain in practice. 
We artificially ablate parts of the data as described in the different experiments to simulate different realistic scenarios of marker and class combinations. 
The fetal liver dataset is separately described in the \textit{Segmentation of fetal liver vasculature} subsection.

\subsection*{Marker combination strategies}\label{sec:methods_ts}
We study herein two fundamentally different settings. First, we study a relatively simpler scenario where training data is acquired with a consistent, fixed set of available markers $m_{1, \dots, K}$; while at test time segmentation is requested for samples only with a subset of trained markers. 
In this setting we ablate subsets of markers to generate test samples with missing markers, i.e.: $m_G \; \forall G \subseteq \{1, \dots, K\}$.
Note that such training data with many markers are scarce as they require extensive workforce, effort, and budget; not only for the sample preparation and acquisitions, but also for their annotations.

Next, we study the more common (and more challenging) scenario where a training set with a fixed set of markers is unavailable.
Since it is computationally not feasible to evaluate each possible set of markers during training, we propose to evaluate a number of illustrative cases (Fig.\,~\ref{fig:f4}). 
To this end, we predefine a set of marker combinations $M^\mathit{train}$ so that specific combinations $m_G^i \in M^\mathit{train}$ are assigned to each sample $s^i$ in the training set.
The test set is constructed with all possible combinations of markers as above, i.e.\ $m_G \; \forall G \subseteq \{1, \dots, K\}$. 
To avoid overfitting to a specific marker combination, we use a validation set with samples that contain all the markers, i.e.\ $m_{1, \dots, K}$. 
In addition,  $M^\mathit{train}$ is shuffled for each cross-validation step so that different $m_G^i$ are applied to different samples.

The resulting segmentation performance understandably varies largely across different marker sets $m_G$. 
Such variation in any metric would be difficult to interpret and to use to compare the models. 
We therefore report relative changes, i.e.\ the score $Q$ achieved by a model $\phi$ on a marker combination $m_G$ relative to a reference model $\phi_\text{ref}$ as $Q_{\phi, \phi_\text{ref}}(m_G) = Q_\phi(m_G) - Q_{\phi_\text{ref}}(m_G)$. 
Consequently, a resulting vector encodes metric differences of a method with respect to a reference, on a given marker combination $m_G$.
To put such results in context, we also report results referenced to an ideal Upper Bound model (\emph{UB}) as $Q_{\phi, \phi_\text{UB}}(m_G) = Q_\phi(m_G) - Q_{\phi_\text{UB}^{m_G}}$, where $\phi_\text{UB}^{m_G}$ is a \emph{UNet} model trained exclusively on $m_G$.
Note that such UB reference results $Q_{\phi_\text{UB}^{m_G}}$ can only be computed for marker combinations explicitly available in at least a sample, i.e.\ for marker combinations
\begin{equation*}\label{eq:markers_ub}
     m_G \in M^\text{UB}, \quad \textrm{where}\quad
     M^\text{UB} = \big\{m_G \, | \, G \subseteq m^i, \; \forall m^i \in  M^\mathit{train} \big\}\ .
\end{equation*}
In contrast, $Q_\phi$ with the proposed models can be computed on any marker combination, i.e.
\begin{equation*}\label{eq:markers_union}
    m_G \in M^\text{total}, \quad \textrm{where}\quad
    M^\text{total} = \big\{m_G \, | \, G \subseteq \bigcup_{m^i \in M^\mathit{train}} m^i \big\}
\end{equation*}
In other words, the proposed methods can produce results even for combinations $m_G$ that never occur in any training sample, as long as the individual markers are present in at least one sample.
Thus, the ratio $|M^\text{UB}| / |M^\text{total}|$ is also reported to indicate the maximum possible marker combinations achievable by conventional networks. 
Lower ratios then emphasize that, besides requiring comprehensive training sets, standard neural network models for segmentation can only be utilized for few marker combinations, while our proposed model can accommodate all combinations.

\subsection*{Network architectures}\label{sec:nn_architecture}
The network architectures employed in this work are described below, first the two baseline models adapted from previous work, followed by our proposed models:
    \begin{itemize}
        \item \emph{UNet} \cite{falk2019u} is one of the most commonly used segmentation models of biomedical images. It is a CNN based on an encoder-decoder architecture with skip connections, targeted to extract features at different resolution levels. We used its standard settings, except for decreasing the number of parameters in each layer by a factor of two, having empirically found that this produces superior results on our dataset. 
        
        \item \emph{HeMIS} \cite{havaei2016hemis} is the current state-of-the-art for image segmentation with missing modalities. Each marker is processed by separate models (\emph{backends}) that are subsequently combined with mean and standard deviation (\emph{abstraction layer}). Such feature aggregation using statistical moments allows to seamlessly apply backpropagation regardless of the missing modalities. The merged output is then processed with additional convolutional layers (\emph{frontend}) to obtain the final segmentation. 

        \item \emph{Marker Zero (MZ)} is a \emph{UNet} where missing markers are padded with zeros, i.e., for any marker combination $m_G$, an input image
        \begin{equation*}\label{eq:mz}
            x_k \leftarrow
            \begin{cases}
                x_k & \text{if } k \in G\\
                0 & \text{otherwise}\ .
            \end{cases}
        \end{equation*}
        
        \item \emph{Marker Sampling (MS)} is a \emph{UNet} prepended with a sampling layer that randomly deletes markers at training time with probability $r_\text{drop} \in (0, 1)$, i.e.\ $x_k \leftarrow \mathit{Bern}(r_\text{drop}) x_k$, where $\mathit{Bern}(r)$ denotes random sampling from a Bernoulli distribution with probability $r$. 
        $r_\text{drop}=0$ would be equivalent to \emph{MZ}, and $r_\text{drop}=1$ is excluded because that would create an input image without markers. 
        Note that no such sampling occurs at inference, but it is implicit to the application strategy in that samples with different markers are expected. 
        We set $r_\text{drop} = 0.5$ in all our experiments, hence all marker combinations are sampled with equal probability.
        In-depth analysis revealed that this choice leads to the best overall accuracy (Supplementary Fig.\,~\ref{fig:sf5}a).
        Note that in this study we consider the scenario where all marker combinations are equally likely to occur at inference, which justifies the choice of setting $r_\text{drop}=0.5$. 
        However, if combinations of specific number of markers are expected to occur more frequently at inference, $r_\text{drop}$ can be adjusted accordingly (Supplementary Fig.\,~\ref{fig:sf5}b), potentially also incorporating marker-specific preferences.
        
        \item \emph{Marker Sampling - Dropout (MS-DR)} is similarly to \emph{MS}, where the sampling layer is replaced with a Dropout layer, i.e.\ $x_k \leftarrow {r^{-1}_\text{drop}} \mathit{Bern}(r_\text{drop}) x_k$ only at training time.
        
        \item \emph{Marker Sampling - Variable Ratio (MS-VR)} is a modification of \emph{MS} where the sampling layer normalizes the output according to the number of sampled markers, i.e.\ $x_k \leftarrow p^{-1} \mathit{Bern}(r_\text{drop}) x_k$ at training time, with $p$ being the ratio of markers chosen by the sampling layer.
        In this case, the normalization $p$ is also applied at inference time, i.e.\ $x_k \leftarrow p^{-1} x_k$, since it it is a function of the number of utilized markers, instead of an expected probability as in \emph{MS-DR}.
        
        \item \emph{HeMIS - Marker Sampling} (\emph{HeMIS-MS}) is a \emph{HeMIS} network that is prepended with our MS sampling layer. In contrast to zeroing out missing markers as in the \emph{UNet} counterpart, we herein ablate any backend corresponding to a missing marker, in order to avoid statistical moment calculations from empty feature maps.
    
        \item \emph{Marker Sampling - Squeeze And Excitation (MS-SE)} is an \emph{MS} model with a \emph{Squeeze and Excitation} (\emph{SE}) module as proposed in \cite{hu2018squeeze} in each \emph{UNet} block (see Fig.\,~\ref{fig:f3}b). 
        These modules each learn a weighting function ${\delta: \mathbb{R}^{1 \times F} \rightarrow \mathbb{R}^{1 \times F}}$ of a feature map $X$ with $F$ activations, where each activation layer $X_{f\in \{1, \dots, F\}}$ is weighted as a function of its layer-wise mean value $\langle X_f \rangle$, i.e.\ $X_f \leftarrow \delta_f(\langle X_f \rangle) X_f$.
        $\delta$ is parametrized using two fully-connected network layers, respectively, with $F/2$ and $F$ nodes and no biases. The first layer is followed by a rectified linear unit (ReLU) and the second one by a sigmoid function.
        
        \item \emph{Marker Sampling - Marker Excite (MS-ME)} instead weights the activations using our proposed \emph{Marker Excite} (\emph{ME}) module (illustrated in Fig.\,~\ref{fig:f3}a) that learns a weighting from a binary vector $V$ encoding which of the $K$ markers are presented to the network as input. 
        Our module thus learns a function ${\omega: \{0, 1\}^{K} \rightarrow \mathbb{R}^{1 \times F}}$ to weight the activations as ${X_f \leftarrow \omega_f(V) X_f}$. Similar to \emph{SE} we parametrize $\omega$ using two fully-connected layers. 
        The first layer has as many nodes as possible marker combinations ($2^K-1$) and we employ biases in both layers, which is seen to increase $F1$-score (Supplementary Fig.\,~\ref{fig:sf4}). 
        Similarly to \emph{MS-SE}, we add \emph{ME} modules after each \emph{UNet} block (Fig.\,~\ref{fig:f3}b).
    \end{itemize}
    
We do not apply Batch Normalization in any of the models as we empirically found this to decrease the $F1$-score, presumably due to the relatively small batch sizes. 
Supplementary Table~\ref{tab:complexity} shows that the \emph{HeMIS} architecture has less parameters than \emph{UNet}, but a higher memory footprint and lower speed. 
Meanwhile, our \emph{MS} and \emph{ME} methods have little to no additional burden, neither in memory space nor in speed. 

\subsection*{Training and evaluation details}\label{sec:nettrain}
We implement all CNN models in TensorFlow 2.1 \cite{tensorflow2015-whitepaper} and train them on an NVIDIA GeForce GTX TITAN X GPU with 12 GB of VRAM, with 3 Intel Xeon E5-2640 v3 CPU cores and 40GB of host RAM. 
Due to large image sizes and GPU memory limitations, we use a rather small batch size of~2. 
Using the Adam optimizer \cite{kingma2014adam}, we minimize the following weighted cross-entropy loss:
\begin{equation*}\label{eq:loss}
    L = \sum\limits_{c \in C} W_c \sum\limits_{y_j^c \in Y^c} y_j^c\log(\overline{y_j^c}),
\end{equation*}
where $y_j^c$ is the ground truth annotation for class $c$, $\overline{y_j^c}$ is its corresponding network prediction, and $W_c$ is a class-specific weight to account for class imbalance. 
We found that using a weight that is linear in the class cardinality leads to training instabilities, especially with large class imbalance. 
Therefore we instead use a logarithmic weighting as follows:
\begin{equation*}\label{eq:classweights}
    W_c = \log \Big(\epsilon + \frac{1}{C |Y^c|} \sum\limits_{i \in \{1, \dots, C\}} |Y^i| \Big),
\end{equation*}
where $|Y^c|$ is the total number of annotated pixels for class $c$ and $\epsilon$ is a small number for stability in logarithm calculation.

The results are evaluated using $F1$-score, individually for each of the classes, as follows:
\begin{equation*}\label{eq:dice}
    F1 = \frac{2TP}{2TP + FP + FN}
\end{equation*}
All networks are trained for 200 epochs, and we choose for evaluation the one that yields the highest mean $F1$-score across all classes on the validation set.
The evolution of the loss function and the $F1$-score is shown in Supplementary Fig.\,~\ref{fig:sf7}.
Note that there exists a considerable class imbalance (see Supplementary Table~\ref{tab:dataset_bm}), which can explain the inferior $F1$-scores when classes are targeted simultaneously (Fig.\,~\ref{fig:f1}b,f) as compared to targeting them individually, which is why we chose the latter approach to conduct our experiments.
There has been substantial research in class imbalance \cite{li2019overfitting}, but this is not within the scope of this paper, where we focus on effects of markers, i.e.\ input image modalities.

The 230 image patches were split according to which of the 8 FM samples in Supplementary Table~\ref{tab:dataset_bm} they belonged to, with 5 samples for training, 1 for validation, and 2 for testing. 
For all experiments we employ 4-fold cross-validation, by ensuring that all samples are once in test set and that the same sample folds are used for different methods in each experimental setting such that their results can be directly compared. 

\subsection*{Tiling strategy for handling large datasets}\label{sec:tiling}
Wide-tissue FM images typically have very large pixel sizes, exceeding memory capacity of typical GPUs used for deep learning. In addition, FM samples are often acquired with different resolutions. 
To address these problems, we herein adopt a pipeline similar to those included in recent \emph{UNet} variants~\cite{falk2019u, gomez2019deepimagej} by decomposing each sample $s^i$ in $J$ patches $p^i_j, \, j \in \{1, \dots, J\}$ with constant size and resolution to segment them individually and to subsequently stitch them together.
These patches are constructed as follows (illustrated in Supplementary Fig.\,~\ref{fig:sf3}):
\begin{enumerate}
    \item We resample the resolution of all FM samples to 1$\mu$m pixel size.
    \item Since convolution operations decrease the spatial extent of the image, we first zero-pad the complete sample with a margin of 92 pixels (half of the difference between input and output patch sizes) required to preserve the original size when stitching the output patches. 
    
    \item The input patches are taken with an overlap of 92 pixels (same as the padding) between them so that there is no overlap in the output patches. In this way, we limit padding artefacts to the border of the samples instead of introducing them for each individual patch used in the CNNs.
    
    \item To normalize appearance differences across samples, for each patch we apply Gaussian standardization (zero mean, unit variance) using the mean and standard deviation of the respective sample that the patch comes from.
    
    \item Output patches are neither overlapping nor padded. Since the sample size does not have to be divisible by the patch size, the last patch of each dimension was taken with as much overlap with the previous one as needed to cover the whole sample. At inference, the result for the overlapping region is averaged for the involved patches. Finally, the slices reconstructed by 2D tiling are inserted in their corresponding axial position to form 3D images. 
\end{enumerate} 

\subsection*{Bone marrow quantification pipeline}\label{sec:methods_bmquant}
For the results reported on the characterization of bone marrow vasculature we employed additional unlabeled samples from the dataset described in~\cite{gomariz2018quantitative}. 
In that earlier work, several samples had to be discarded because their image quality was not sufficient for the simplistic image processing tools (MIP) employed therein. 

The \emph{MS-ME} model proposed herein allows for processing of samples with diverse marker combinations. 
However, as shown in Fig.\,~\ref{fig:f1}c-e\&g-i, not all markers are beneficial for achieving a precise segmentation. 
In order to provide an accurate quantification without sacrificing many samples, we employ in our analysis only samples that are stained with the marker sets in $M_2^s$ for sinusoids and in $M_2^a$ for arteries, as shown respectively in Fig.\,~\ref{fig:f1}c and Fig.\,~\ref{fig:f1}g. 
Quantified on the samples stained with the best marker combination, the above marker sets guarantee an $F1$-score of 75\% or higher than that achievable given the best marker combination. 
With this criterion, 47 samples are employed for the quantification of sinusoids and 29 for arteries. 

All employed samples were visually inspected to qualitatively confirm that the segmentation was satisfactory (examples in Supplementary Fig.\,~\ref{fig:sf1}).
The $F1$-score reported for \emph{MS-ME} in Table~\ref{tab:bmquant} is calculated from the results reported in Fig.\,~\ref{fig:f3}e aggregated over the markers in $M_2^s$ for sinusoids and in $M_2^a$ for arteries. 
The \emph{MIP} method was evaluated on the same annotations as the other methods proposed in this work (Supplementary Table~\ref{tab:dataset_bm}), although not all of these samples were employed for the quantification in~\cite{gomariz2018quantitative}.

FM images often contain out of tissue regions which must be discarded for analysis purposes. To this end, we account in this analysis for an extra class denoted as \emph{tissue}.
In several works, segmentation of this class is easily achieved by thresholding of the marker $m_1$ (DAPI) and some simple morphological image processing \cite{gomariz2019imaging}. 
Given the simplicity of the task, we train a separate \emph{UNet} model that uses only $m_1$ (available in all the samples) and achieves an $F1$-score of $88.7\pm6.8$ with the same evaluation strategy applied for other models in this work. 
The reported fraction of bone marrow volume for sinusoids and arteries is calculated as the ratio of their respective foreground pixels within the newly defined tissue class to the total number of foreground tissue pixels. 

\subsection*{Segmentation of fetal liver vasculature}
For the fetal liver dataset presented in Results section \emph{Marker Sampling and Excite for the quantification of fetal liver vasculature}, we employed the same strategies previously described for the earlier bone marrow dataset.
We herein report the dataset description and technical details that differ from the bone marrow images. 

We denote the 5 available markers with letters instead of numbers (i.e.\ the complete marker set is denoted as $m_{abcde}$): \emph{DAPI}~($a$), \emph{lyve1}~($b$), \emph{Hlf}~($c$), \emph{Evi1} or \emph{$\alpha$-catulin}~($d$), and smooth muscle actin~($e$). 
The class to segment in this dataset is denoted as \emph{vessels}.
The number of samples, patches, and percentage of annotated pixels can be seen in Supplementary Table~\ref{tab:dataset_fl}.

The vasculature quantification for different embryonic stages denoted as E13.5 and E17.5 displayed in Fig.\,~\ref{fig:f6}g is performed on 15 unlabeled samples: 8 samples of E13.5 and 7 samples of E17.5.
A manually annotated 3D \emph{tissue mask} is included in these images that defines the total tissue volume, which is employed to disregard out-of-tissue regions in the analysis of vasculature occupation. 
Note that that the annotated fetal liver vasculature data employed for training and evaluation of segmentation CNNs is a subset of these samples. 

\subsection*{Statistical tests}
Unless otherwise specified, the two-sided Wilcoxon signed-rank test was employed to assess differences between paired test results, which is non-parametric to avoid any assumption of normal distribution. 
Since the aim of the models was to perform best across different classes, the employed tests aggregated the data for both sinusoids and arteries in a paired-wise manner.  
When the data under evaluation was unpaired, we used two-sided Mann-Whitney U test, which is also non-parametric. 
Unless otherwise stated, measurements were always taken from distinct samples.
Significance was established with a p-value$\leq$0.05.
Boxplots employed in the different figures consist of a box that represents the data quartiles and whiskers that indicate the extent of data points within 1.5 times the interquartile range. 

\subsection*{Animal studies}
Mice C57BL/6J were purchased from Charlesriver. 
For bone marrow studies, mice were analysed at 12–20 weeks of age. For fetal liver analyses, embryos were extracted from the previously euthanized pregnant dams at the developmental stages indicated, through a small abdominal incision and further dissected with forceps under a stereo microscope, where single lobes were separated using a surgical scalpel.
The lobes were fixed in 2\% paraformaldehyde (diluted in PBS) (6h, 4\textdegree C), washed twice in ice-cold PBS for five minutes and subsequently dehydrated in 30\% sucrose in PBS (24–48h, 4\textdegree C). The liver lobes were placed into base molds (15×15×5 mm) and completely covered in OCT medium, snap-frozen using liquid nitrogen and stored at -80\textdegree C until use. Slices were stained following the same protocols described in~\cite{gomariz2018quantitative} for marrow tissues.
Experimental animals were not randomized and experiments were performed in a non-blinded fashion. 
Animals were maintained at the animal facility of the University Hospital Zürich and treated in accordance with the guidelines of the Swiss Federal Veterinary Office. 
Experiments and procedures were approved by the Veterin\"aramt des Kantons Z\"urich, Switzerland.

\section*{Data availability}
The labeled dataset employed for training and evaluation of the models described is included as a single HDF5 file within a CodeOcean capsule in \url{https://codeocean.com/capsule/8424915/tree/v1}\cite{MSME_CO}. 

\section*{Code availability}
The code employed for training the models described in this paper is publicly available on the CodeOcean platform as \url{https://codeocean.com/capsule/8424915/tree/v1}\cite{MSME_CO}. 
This capsule also includes the trained models employed for the different presented figures. 
\emph{MS-ME} is also implemented within \emph{MiNTiF}, our Fiji plugin for ImageJ for  user-friendly training and deployment of CNNs by non-experts of deep learning: \url{https://github.com/CAiM-lab/MiNTiF}.

\section*{Acknowledgments}
We thank Takashi Nagasawa for the Cxcl12-GFP reporter mice, Tomomasa Yokomizo for the Hlf-tdTomato mouse strain and Mineo Kurokawa for the Evi1-GFP mouse strain. 
This work was enabled by funding from the Hasler Foundation (A.G.,\ O.G.).
We also acknowledge the support from the Swiss National Science Foundation (SNSF) 179116 (O.G.) \& 310030\_185171 (C.N.-A.), the Swiss Cancer Research Foundation KFS-3986-08-2016 (C.N.-A.), the Clinical Research Priority Program ``ImmunoCure'' of the University of Zurich to (C.N.-A.).
We extend our thanks to NVIDIA for their GPU support.

\section*{Author contributions}
A.G.\ designed and performed the experiments and drafted the manuscript.
T.P.\ and O.G.\ provided key methodological ideas and input.
A.G., P.M.H., S.I.\ and U.S.\ conducted data acquisition, manual labeling of the images, and helped interpret bone marrow analysis in the biological context.
T.P., O.G., and C.N.-A.\ critically revised the text. 
A.G., C.N.-A., and O.G.\ conceived this project.
This work was jointly directed by C.N.-A.\ for the biological and data acquisition aspects and O.G.\ for computer scientific and methodological aspects. 
All authors discussed the results and provided feedback to the writing.

\newpage
\section*{Supplementary Materials}

\begin{suppfigure}[hbt!]
    \centering
    \includegraphics[width=0.9\textwidth]{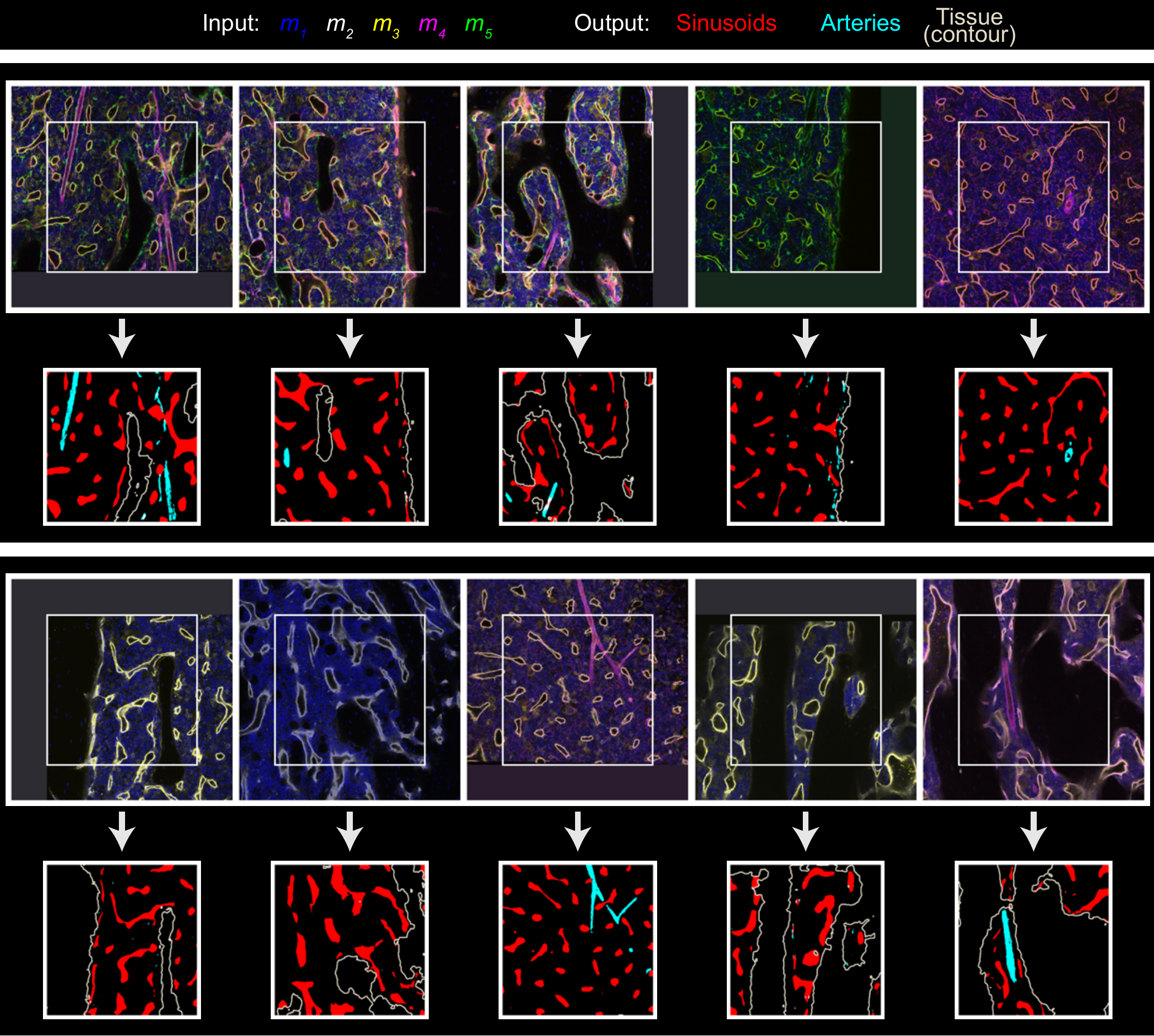}
    \caption{Example images for the qualitative assessment of the segmentation of bone marrow images employed for the quantification of vasculature. Input images contain different combinations of markers shown as an overlay of different colors. The white rectangle within the input images represents the size of the output image when processing with a neural network. White arrows depict inference with the \emph{MS-ME} model. Different colors are employed in the output images to show the different predicted classes. Since the tissue class overlaps with the other two, its contour is used instead for visual purposes. }
    \label{fig:sf1}
\end{suppfigure}

\begin{suppfigure}[hbt!]
    \centering
    \includegraphics[width=0.25\textwidth]{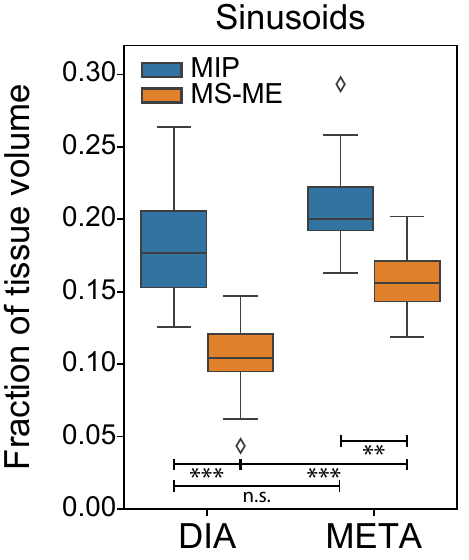}
    \caption{Comparison of bone marrow volume ratio occupied by sinusoids in both diaphysis (DIA) and metaphysis (META) when segmenting them with the morphological image processing (MIP) algorithm previously proposed (n=12 for both DIA and META) and with our MS-ME method proposed here (n=61 for DIA, n=24 for META).}
    \label{fig:sf2}
\end{suppfigure}

\begin{suppfigure}[hbt!]
    \centering
    \includegraphics[width=0.9\textwidth]{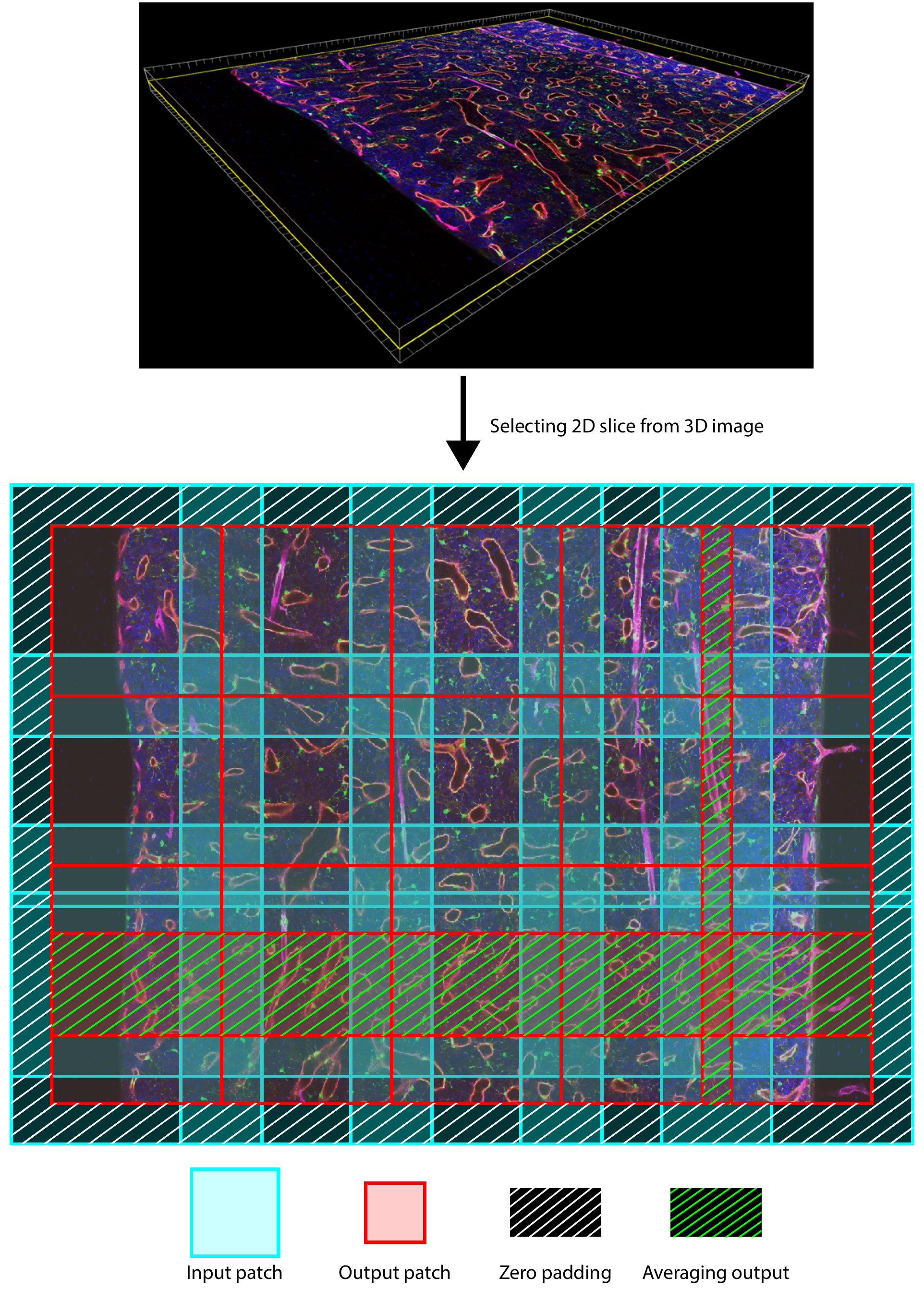}
    \caption{Illustration of the image tiling pipeline employed to create suitable patches for CNNs. 
    An example of a slice within the 3D image frame is shown in the upper part using Imaris (Bitplane AG). 
    That slice is decomposed in patches as illustrated in the lower part. Each output patch (red) is smaller than their corresponding input patch (cyan) due to the convolutional operations in CNNs. We position the output patches next to each other without overlap in order to avoid padding artifacts in the application of CNNs. Instead, zero padding is only applied along the borders of the whole slice (area with white stripes). When an overlap between output patches cannot be avoided to fill the slice (area with green stripes), the average of the different patches in that region is used. 
    }
    \label{fig:sf3}
\end{suppfigure}

\begin{suppfigure}[hbt!]
    \centering
    \includegraphics[width=0.15\textwidth]{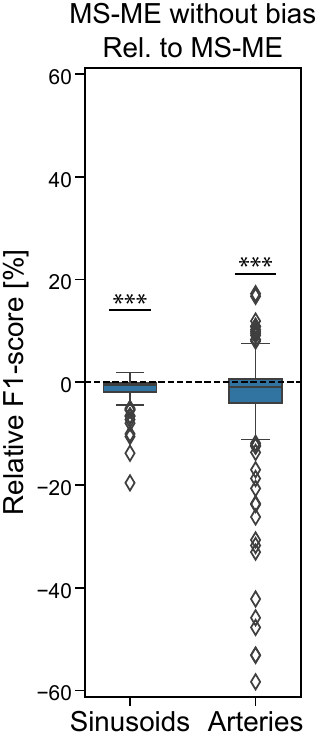}
    \caption{
    Effect of bias term on \emph{ME} module. $F1$-score of \emph{MS-ME} model where the bias terms for all \emph{ME} modules have been removed, relative to the proposed \emph{MS-ME} model with bias across all marker combinations and cross-validation steps. 
    }
    \label{fig:sf4}
\end{suppfigure}

\begin{suppfigure}[hbt!]
    \centering
    \includegraphics[width=0.85\textwidth]{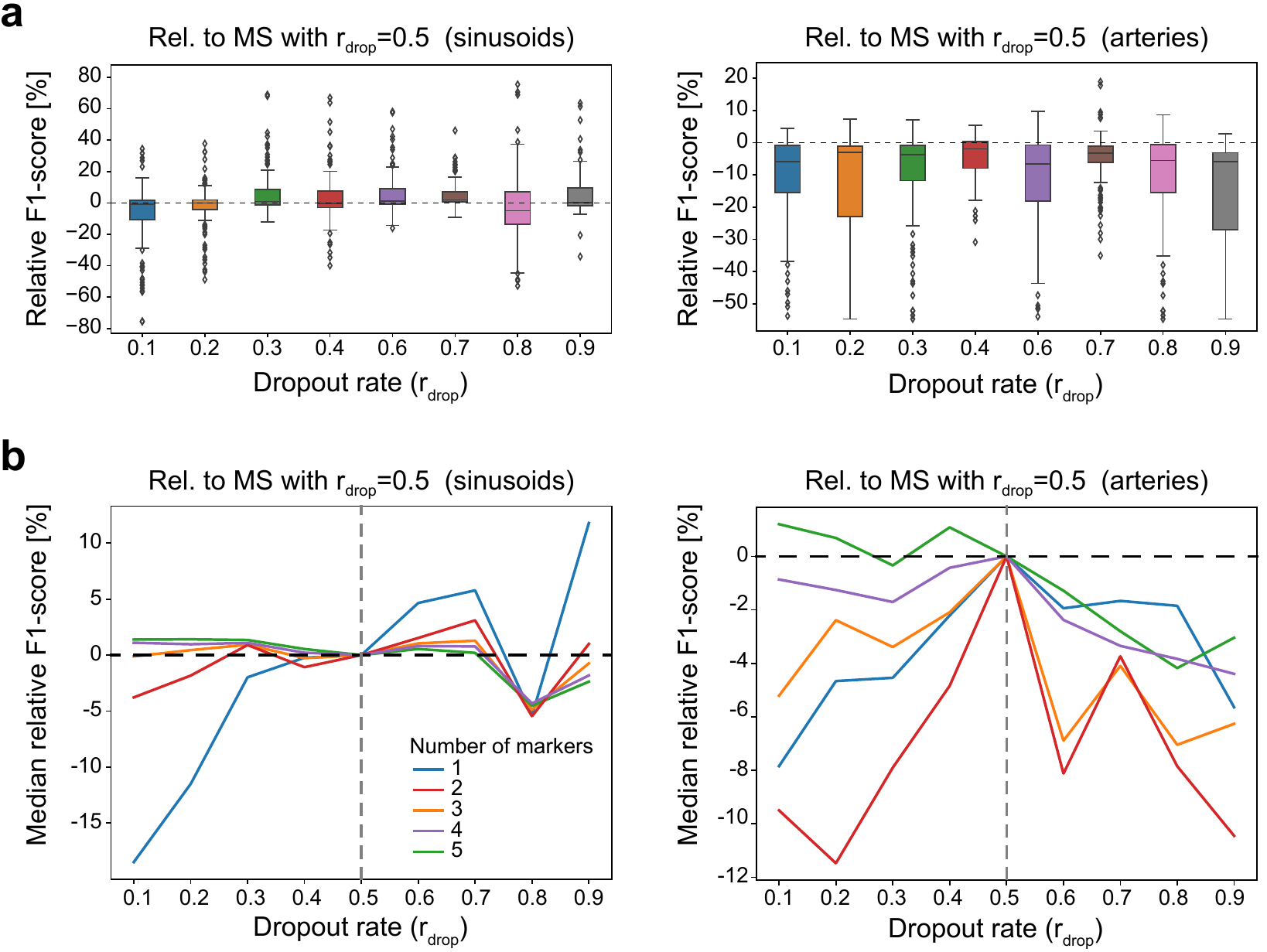}
    \caption{
    Effect of marker dropout rate $r_\textrm{drop}$ in \emph{MS}. 
    $F1$-score of \emph{MS} models with different $r_\textrm{drop}$ evaluated on the sinusoids (left) and arteries (right) relative to the proposed \emph{MS} with $r_\textrm{drop}=0.5$.
    (\textbf{a}) Evaluation for all 31 possible marker combinations (n=124).
    Whereas some $r_\textrm{drop}\neq 0.5$ produce a slightly superior $F1$-score for sinusoids, $r_\textrm{drop}=0.5$ is the best option for arteries and overall. 
    (\textbf{b}) Median relative $F1$-score for models evaluated on combinations of specific numbers of markers, each represented by a different color for the different $r_\textrm{drop}$ ($n=\frac{K!}{(K-k)!k!}$, where $K$ is the number of markers available, and $k$ the number of markers considered for each evaluation).
    Smaller $r_\textrm{drop}$ are shown to be beneficial for combinations of more markers, and vice-versa.
    However, this trend becomes noisier for $r_\textrm{drop}>0.5$, as illustrated with the gray dashed line. 
    This effect can be due to the decrease in markers observed over time, although it is a question worth of further investigation in future work. 
    }
    \label{fig:sf5}
\end{suppfigure}

\begin{suppfigure}[hbt!]
    \centering
    \includegraphics[width=0.85\textwidth]{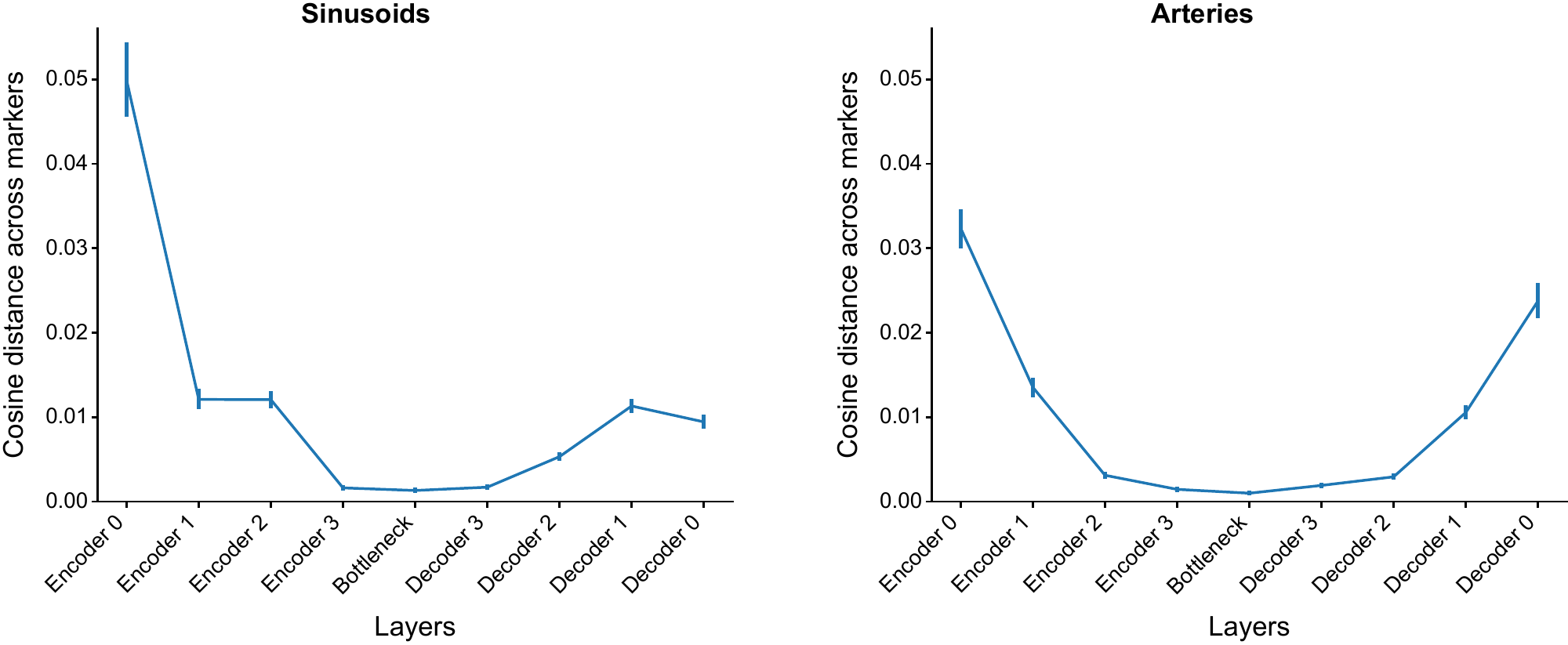}
    \caption{
    Analysis of attention parameters in \emph{ME} modules for sinusoids (left) and arteries (right).
    We estimate recalibration strength by calculating cosine distances between the \emph{ME} attention subnetwork outputs obtained for each of the possible input marker combinations. 
    Results are represented as the mean of all such pair-wise distances between all possible marker combinations, at a given layer where \emph{ME} is placed, with the bars depicting the standard deviations of these distances. 
    Using the colored network layers shown in Fig.\,~\ref{fig:f3}b, \emph{Encoder} layers correspond to the network layers in blue, \emph{Decoder} to the layers in green, and \emph{Bottleneck} to the yellow layers.
    The numbers next to each layer indicate the resolution level, where 0 corresponds to the highest (original resolution) and 3 to the lowest (i.e., right before and after the bottleneck, for the encoder and decoder, respectively). 
    It can be seen in this representation that recalibration strength is higher in layers with higher resolution, especially near the input of the network.
    This observation may indicate that high resolution layers of the network focus on effectively combining features from available markers, and in this way create shared abstract features that are common across markers for subsequent processing in lower resolution layers.
    }
    \label{fig:sf6}
\end{suppfigure}

\begin{suppfigure}[hbt!]
    \centering
    \includegraphics[width=0.75\textwidth]{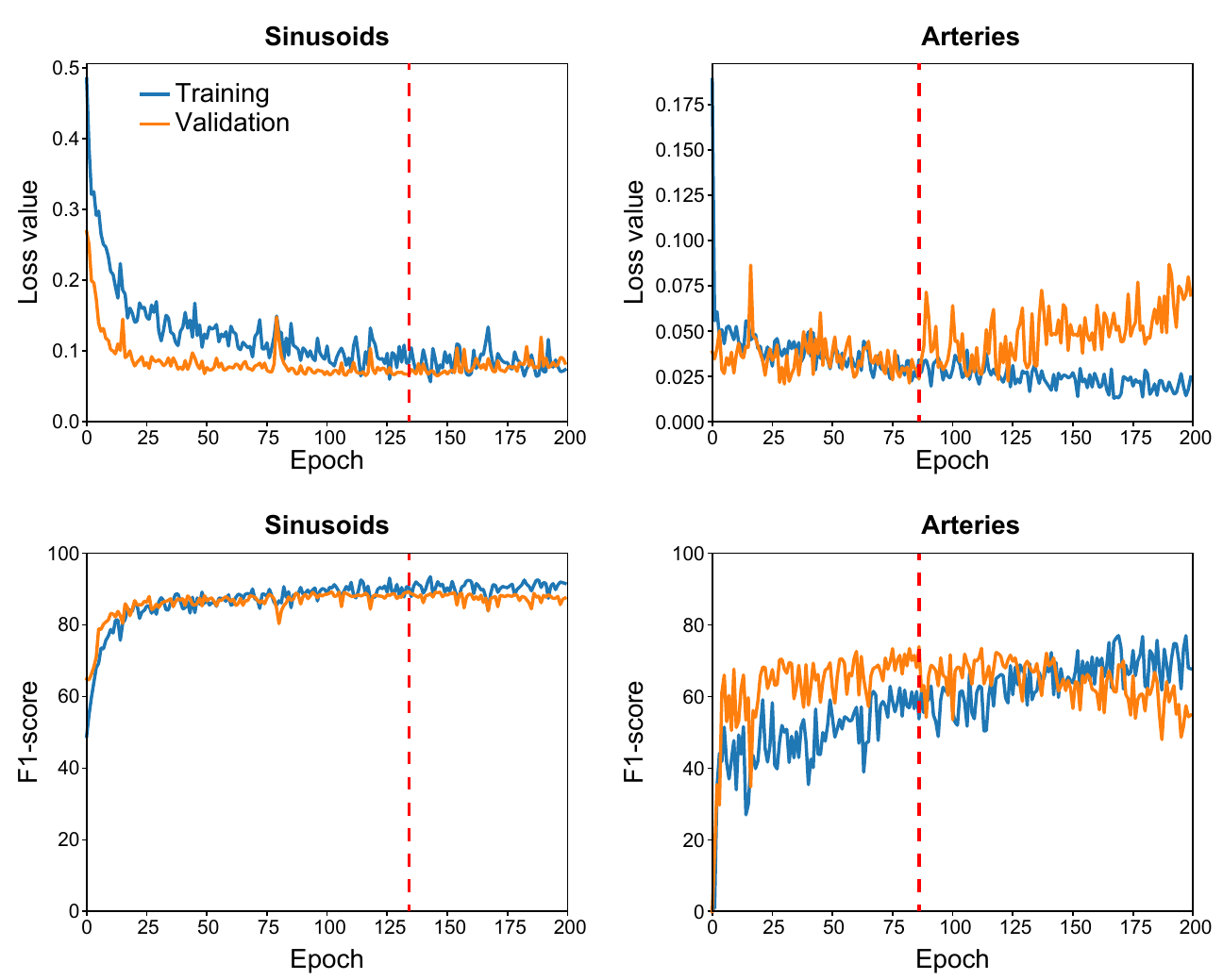}
    \caption{
    Evolution of the weighted cross-entropy loss (top) and the $F1$-score (bottom) across epochs for the training (blue) and validation (orange) sets with our proposed \emph{MS-ME} model, both for models trained for segmentation of sinusoids (left) and arteries (right). 
    The red dashed line marks the epoch at which we choose the model, based on the highest validation $F1$-score.
    }
    \label{fig:sf7}
\end{suppfigure}

\begin{supptable}[hbt!]
    \centering
    \scalebox{0.8}{
    \begin{tabular}{c|ccc}
    \textbf{sample id ($s^i$)} & \textbf{\# patches ($J^i$)} & \textbf{sinusoids (\%)} & \textbf{arteries (\%)} \\
    \hline
    1 & 54 & 9.71 & 0.29 \\
    2 & 12 & 7.87 & 0.99 \\
    3 & 30 & 11.97 & 0.21 \\
    4 & 12 & 11.20 & 0.98 \\
    5 & 12 & 10.93 & 0.58 \\
    6 & 48 & 13.60 & 0.24 \\
    7 & 54 & 11.23 & 0.24 \\
    8 & 8 & 15.19 & 0.27 \\
    TOTAL & 230 & 11.41 & 0.34 \\
    \end{tabular}
    }
    \caption{Summary of the annotated bone marrow dataset employed for training and evaluation of the proposed methods. There are 8 different samples composed of different annotated patches. The annotated classes are either sinusoids or arteries. The remaining pixels are considered background. There is a notable class imbalance, with most of the pixels being background and arteries being a minority class.  }
    \label{tab:dataset_bm}
\end{supptable}

\begin{supptable}[hbt!]
    \centering
    \scalebox{0.8}{
    \begin{tabular}{c|cc}
    \textbf{sample id ($s^i$)} & \textbf{\# patches ($J^i$)} & \textbf{vessels (\%)}\\
    \hline
    0 & 4 & 20.33 \\
    1 & 8 & 20.81 \\
    2 & 4 & 18.88 \\
    3 & 4 & 13.25 \\
    4 & 4 & 20.54 \\
    5 & 8 & 21.11 \\
    6 & 4 & 25.43 \\
    7 & 4 & 15.41 \\
    8 & 4 & 27.20 \\
    9 & 4 & 10.04 \\
    10 & 8 & 12.78 \\
    11 & 4 & 25.04 \\
    12 & 4 & 17.90 \\
    13 & 4 & 21.67 \\
    14 & 4 & 31.79 \\
    15 & 4 & 5.32 \\
    16 & 8 & 6.97 \\
    17 & 4 & 23.02 \\
    18 & 9 & 18.69 \\
    19 & 4 & 18.39 \\
    20 & 4 & 20.79 \\
    21 & 4 & 27.87 \\
    22 & 4 & 22.87 \\
    23 & 8 & 12.27 \\
    24 & 4 & 1.41 \\
    25 & 17 & 16.97 \\
    TOTAL & 142 & 17.72 \\
    \end{tabular}
    }
    \caption{
    Summary of the annotated dataset employed for the segmentation of fetal liver vasculature, where the only class is named \emph{vessels} and the remaining pixels are considered background. 
    }
    \label{tab:dataset_fl}
\end{supptable}

\begin{supptable}[hbt!]
    \centering
    \scalebox{0.8}{
    \begin{tabular}{l|cccc}
    \textbf{Models} & \textbf{\makecell{Training time \\(ms/batch)}} & \textbf{\makecell{Inference time \\(ms/batch)}} & \textbf{\makecell{GPU memory \\footprint (GB)}} & \textbf{\makecell{\# parameters\\(x10$^6$)}} \\
    \hline
    MZ & 0.28 & 0.20 & 0.95 & 7.76 \\
    HeMIS & 1.07 & 0.17 & 3.8 & 0.35 \\
    MS & 0.32 & 0.19 & 0.95 & 7.76 \\
    HeMIS-MS & 1.00 & 0.20 & 3.9 & 0.35 \\
    MS-DR & 0.34 & 0.17 & 0.95 & 7.76 \\
    MS-VR & 0.32 & 0.18 & 0.95 & 7.76 \\
    MS-SE & 0.35 & 0.15 & 1.12 & 8.2 \\
    MS-ME & 0.34 & 0.2 & 1.16 & 7.81 \\
    MS$+$ & 0.25 & 0.18 & 1.05 & 9.35 \\
    UB & 7.57 & 0.2 & 0.95 & 240.56 \\
    \end{tabular}
    }
    \caption{
    Computational complexity of the different models employed. 
    Time is calculated as the median across batches. 
    GPU memory footprint refers to the maximum usage recorded during training. 
    Note that \emph{UB} is an ensemble of 31 models, and hence the increase in parameters only affects the training time.
    A single model is employed at inference for the pertinent marker combination, making the inference time equivalent to that of a single model. 
    }
    \label{tab:complexity}
\end{supptable}

\end{document}